\documentclass[journal]{IEEEtran}
\RequirePackage[T1]{fontenc}
\usepackage{standalone}
\usepackage{cite}
\usepackage{soul}
\usepackage{amsmath,amsfonts,amsthm}
\usepackage{dsfont}
\usepackage{siunitx}
\usepackage{graphicx}
\usepackage[caption=false,font=footnotesize,labelfont=sf,textfont=sf]{subfig}
\usepackage{tikz}
\usepackage{style/catppuccinpalette}
\usetikzlibrary{positioning, arrows.meta, calc, fit}
\usepackage{pgfplots}
\pgfplotsset{compat=newest}
\usepackage{tikzexternal}
\tikzsetexternalprefix{tikz/}
\usepackage{array}
\usepackage{tabularx}
\usepackage{booktabs}
\usepackage{multirow}
\usepackage[normal, flushleft]{threeparttable}
\usepackage{algpseudocode}
\usepackage{algorithm}
\usepackage{xcolor}
\usepackage{hyperref}
\usepackage{style/macros}
\usepackage{style/catppuccinpalette}

\newcolumntype{Y}{>{\centering\arraybackslash}X}

\DeclareMathOperator{\APE}{APE}
\DeclareMathOperator{\mlp}{MLP}
\DeclareMathOperator{\argmin}{argmin}

\newcommand{\observationSpace}{\mathbb{O}}
\newcommand{\baseWavelength}{\lambda_{K}}
\newcommand{\norm}[1]{\left\lVert#1\right\rVert}

\newcommand{\perceptionModel}{\text{PerceptionModel}}

\def\figwidth{0.95}
\def\figwidthDouble{0.97}
\def\tableWidth{1.0}

\definecolor{mLinkBlue}{HTML}{002d7d}
\hypersetup{
	colorlinks,
	linkcolor={mLinkBlue},
	citecolor={mLinkBlue},
	urlcolor={mLinkBlue}
}

\usepackage[capitalize,nameinlink]{cleveref}
\newcommand{\fgref}[1]{\cref{#1}}
\newcommand{\scref}[1]{\Cref{#1}}
\renewcommand{\eqref}[1]{\labelcref{#1}}
\newcommand{\saurav}[1]{}
\newcommand{\damian}[1]{}
\newcommand{\fred}[1]{}

\usepackage{siunitx}
\newcommand*{\SIvel}[1]{\SI{#1}{\meter\per\second}}

\newcommand*{\SIs}[1]{\SI{#1}{\second}}
\newcommand*{\SIm}[1]{\SI{#1}{\meter}}
\newcommand*{\SIkm}[1]{\SI{#1}{\kilo\meter}}

\begin{document}
\bstctlcite{IEEEexample:BSTcontrol}
\title{MAST: Multi-Agent Spatial Transformer for Learning to Collaborate}
\author{Damian Owerko, Frederic Vatnsdal, Saurav Agarwal, Vijay Kumar, Alejandro Ribeiro}

\maketitle

\begin{abstract}
This article presents a novel multi-agent spatial transformer (MAST) for learning communication policies in large-scale decentralized and collaborative multi-robot systems (DC-MRS).
Challenges in collaboration in DC-MRS arise from:
\emph{(i)} partial observable states as robots make only localized perception,
\emph{(ii)} limited communication range with no central server, and
\emph{(iii)} independent execution of actions.
The robots need to optimize a common task-specific objective, which, under the restricted setting, must be done using a communication policy that exhibits the desired collaborative behavior.
The proposed MAST is a decentralized transformer architecture that learns communication policies to compute abstract information to be shared with other agents and processes the received information with the robot's own observations.
The MAST extends the standard transformer with new positional encoding strategies and attention operation that employs windowing to limit the receptive field for MRS.
These are designed for local computation, shift-equivariance, and permutation equivariance, making it a promising approach for DC-MRS.
We demonstrate the efficacy of MAST on decentralized assignment and navigation (DAN) and decentralized coverage control.
Efficiently trained using imitation learning in a centralized setting, the decentralized MAST policy is robust to communication delays, scales to large teams, and performs better than the baselines and other learning-based approaches.
\end{abstract}


\section{Introduction}
Collaborative multi-robot systems enable the execution of tasks that would otherwise be infeasible or inefficient with a single robot in large environments and time-critical applications.
Multi-robot systems, in particular uncrewed aerial vehicles and ground mobile robots, have been deployed in a wide range of tasks including warehouse automation~\cite{Kegeleirs21-SwarmSLAM}, environmental monitoring~\cite{Lu25-Reinforcement}, search and rescue~\cite{Cao24-HMASARMultiAgentSearch}, and exploration~\cite{Bi24-CUREHierarchical}.
Mobile robots are constrained by size, weight, and power (SWaP), which restricts the computational, sensing, and communication capabilities.
Decentralized systems, where individual robots act autonomously using sensed local information and the information received through limited communication with nearby robots, offer scalable and resilient solutions for such SWaP-constrained multi-robot systems.

\begin{figure}[htbp]
    \centering
    \usetikzlibrary{arrows.meta, positioning, calc, fit}

\begin{tikzpicture}[
    x=3em,
    y=3em,
    module/.style={draw, very thick, rectangle, rounded corners, minimum width=\moduleWidth, minimum height=2em, align=center, node distance=\moduleDistance},
    modulesharp/.style={module, rounded corners = 0cm},
    submodule/.style={module, node distance=\submoduleDistance},
    perception/.style={module, rounded corners=0cm, fill=yellow},
    pm/.style={module, fill=red},
    communication/.style={module, rounded corners=0cm, fill=blue},
    projection/.style={submodule, minimum width=\projectionWidth, fill=green},
    attention/.style={submodule, fill=orange},
    mlp/.style={module, fill=orange},
    readout/.style={module, fill=red},
    operator/.style={draw, thick, circle, inner sep=1pt, minimum size=\operatorSize, node distance=\submoduleDistance},
    label/.style={font=\footnotesize},
    edge/.style={thick,rounded corners=0.3em},
    arrow/.style={-{Latex[length=0.5em]}, edge},
    ]
    \def\alpha{80}
    \colorlet{red}{CtpMochaRed!\alpha}
    \colorlet{green}{CtpMochaGreen!\alpha}
    \colorlet{blue}{CtpMochaBlue!\alpha}
    \colorlet{yellow}{CtpMochaYellow!\alpha}
    \colorlet{orange}{CtpMochaPeach!\alpha}

    \def\moduleWidth{12em}
    \def\moduleDistance{2em}
    \def\submoduleDistance{1em}
    \def\operatorSize{1.0em}
    \pgfmathsetmacro{\projectionWidth}{(\moduleWidth - 2*\submoduleDistance)/3}

    \pgfdeclarelayer{background}
    \pgfsetlayers{background,main}

    \coordinate (ColA) at (-0.6*\moduleWidth, -2.5em);
    \coordinate (ColB) at (-0.2*\moduleWidth, -2.5em);
    \coordinate (ColMid) at (0, -2.5em);
    \coordinate (ColC) at (0.2*\moduleWidth, -2.5em);
    \coordinate (ColD) at (0.6*\moduleWidth, -2.5em);

    \node[] (env) at (0,0) {Environment};
    \node[modulesharp,fill=yellow,below=\moduleDistance of env] (pcpn) {Local Sensing};
    \draw[arrow] (env) -- (pcpn);

    \node[module, fill=red, below=\moduleDistance of pcpn] (pm) {Perception Model};
    \node[modulesharp, fill=blue, below=\moduleDistance of pm] (comm) {Communication};

    \draw[arrow] (pcpn) -- node[label, anchor=west] {$\bfo_i(t)$} (pm);
    \draw[arrow] (pm) -- node[label, anchor=west] {$\bfx_i(t)$} (comm);
    \coordinate[below=\moduleDistance of comm] (commout);
    \draw[edge] (comm) -- node[label, pos=0.5, anchor=west] {$\bfX, \bfP$} (commout);

    \coordinate[below=0.5em of commout] (commbranch);
    \node[projection, below=\submoduleDistance of commbranch] (fk) {$\bfr \odot \bfK$};
    \node[projection, left=of fk] (fq) {$\bfr \odot \bfQ$};
    \node[projection, right=of fk] (fv) {$\bfV$};
    \draw[arrow] (commout) -- (commbranch) -| (fk);
    \draw[arrow] (commout) -- (commbranch) -| (fq);
    \draw[arrow] (commout) -- (commbranch) -| (fv);

    \node[submodule, fill=orange, below=of fk] (attn) {Spatial Self-Attention};
    \node[operator, below=of attn] (AddNorm1) {$+$};
    \coordinate[below=0.5em of AddNorm1] (Residual2Start);
    \node[mlp, submodule, below=of Residual2Start] (MLP) {MLP};
    \node[operator, below=of MLP] (AddNorm2) {$+$};

    \draw[arrow] (attn) -- (AddNorm1);
    \draw[arrow] (AddNorm1) -- (MLP);
    \draw[arrow] (MLP) -- (AddNorm2);
    \draw[arrow] (fq) -- (fq|-attn.north);
    \draw[arrow] (fk) -- (fk|-attn.north);
    \draw[arrow] (fv) -- (fv|-attn.north);

    \coordinate (Residual1Corner) at (commbranch-|ColA);
    \draw[arrow] (commout) -- (commbranch) -- (Residual1Corner) |- (AddNorm1);
    \coordinate (Residual2Corner) at (Residual2Start-|ColA);
    \draw[arrow] (Residual2Start) -- (Residual2Corner) |- (AddNorm2);
    \begin{pgfonlayer}{background}
        \node[fit=(fk) (fq) (fv) (commout) (attn) (Residual1Corner) (Residual2Corner) (MLP) (AddNorm1) (AddNorm2), module] (Transformer) {};
    \end{pgfonlayer}
    \node[anchor=east, align=right] at (Transformer.west) {MAST};
    \node[anchor=west, align=right] at (Transformer.east) {$\times L$};

    \draw[arrow] (ColA) node[label, anchor=south] {$\bfp_k(t)$} |- (comm);

    \node[readout, module, below=of AddNorm2] (Readout) {MLP};
    \draw[arrow] (AddNorm2) -- node[label, pos=0.6, anchor=west] {$\bfY$} (Readout);
    \draw[arrow] (Readout.south) -- +(0, -1.2em) node[label, anchor=west] {$\bfu_i(t)$};

\end{tikzpicture}
    \caption{%
      The multi-agent spatial transformer (MAST) architecture within the learnable perception-action-communication (LPAC~\cite{Agarwal2024}) system from the perspective of a robot~\(i\).
        Each agent computes in parallel.
        The $i$th agent is at position $\bfp_i(t)$ and uses its sensing capabilities to make observations $\bfo_i(t)$ of the environment.
        The local perception model transforms those into $\bfx_i(t)$. 
        Each agent communicates its position $\bfp_i(t)$ and embedding $\bfx_i(t)$ to other agents. 
        The $i$th agent collects the received embeddings and positions into $\bfX_i \in \reals^{N \times d}$ and $\bfP_i \in \reals^{N \times 2}$, respectively.  
        MAST encodes positional information into the received embeddings and processes them using $L$ self-attention layers interleaved with MLPs and residual connections. 
        The final self-attention layer outputs a matrix $\bfY_i \in \reals^{N \times d}$.
        Finally, each agent locally projects its row from $\bfY_i$ to the action space $\mathbb{U}$ via an MLP to produce the control $\bfu_i(t)$.%
    \label{fig:architecture-abstract}}
\end{figure}
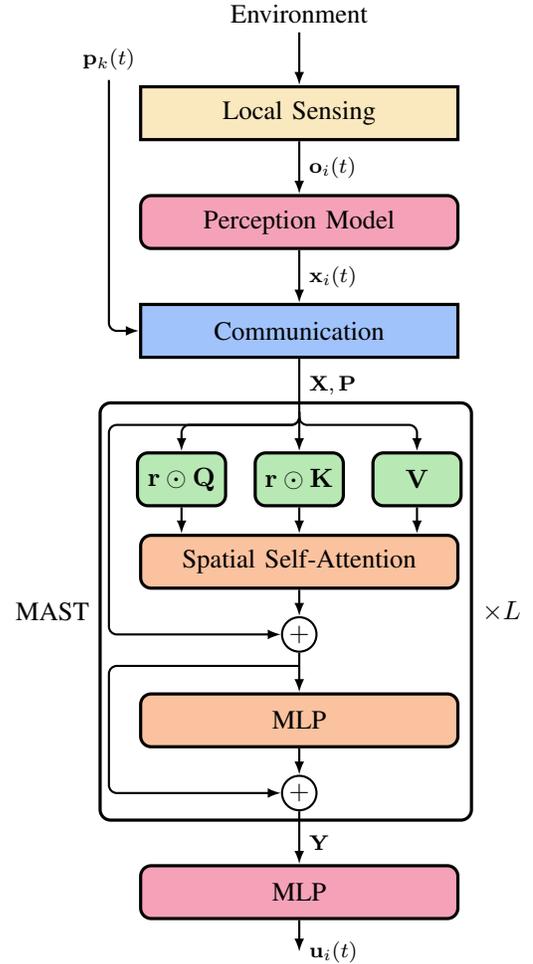

In decentralized systems, no robot has the cumulative state of the entire system.
Any solution must provide each robot with the following two capabilities~\cite{Agarwal2024}:
\emph{(i)}~to determine \emph{what} messages to communicate with nearby agents, and
\emph{(ii)}~\emph{how} to take actions by incorporating the information received from other agents with one's observations.
Ensuring application-specific emergence of effective collaboration through individual actions is fundamentally challenging due to the decentralized nature and limited knowledge of the system's state.
This article presents a novel multi-agent spatial transformer neural network architecture for addressing the problem of emergent collaboration in decentralized multi-robot systems.

Generally, a robot in a decentralized system executes a \emph{perception-action-communication} (PAC) loop.
Based on this, a new paradigm of learning-based approaches has emerged, introducing learnable perception-action communication (LPAC) loops~\cite{Hu22-Scalable, Agarwal2024}.
It breaks down the architectures into three components.
A perception module locally processes inputs from sensors, such as cameras, and any other available information.
Then, a communication module combines the processed information from nearby robots.
Finally, an action module computes a control action from the received information.
The perception and action models are executed independently by each agent.
The communication module is responsible for inter-robot collaboration; therefore, its design is critical for the scalability of a multi-agent system.
The proposed approach in this article utilizes a PAC loop and provides a transformer-based solution for the communication module.

The \emph{primary contribution} of this article is the novel multi-agent spatial transformer (MAST) architecture, which leverages spatial information modeled by the relative positions of nearby agents.
The MAST is an extension of the original transformer architecture for modeling and decentralized execution for large-scale multi-robot systems.
The article:
\emph{(i)}~Introduces a novel positional encoding strategy, designed explicitly for multi-robot systems;
\emph{(ii)}~Empirically demonstrates that by adding appropriate windowing, MAST can generalize to larger teams after training on medium-sized teams; and
\emph{(iii)}~Demonstrates the consistency in decentralized execution of MAST under communication delays.
We support these claims by numerical experiments on two tasks: decentralized assignment and navigation (DAN) and decentralized coverage control.

\section{Related Work}
We split the discussion of related works into three areas: graph neural networks in multi-agent systems, recent applications of transformers to multi-agent systems, and a brief overview of positional encodings for transformers.
We refer the reader to \scref{sec:dan,sec:coverage_control} for an overview of prior works in DAN and coverage control, respectively.

\subsection{Graph Neural Networks in Multi-Agent Systems}
Graph neural networks (GNNs) are a popular backbone architecture for the communication module.
For example, GNNs are effective decentralized controllers in a variety of tasks, including multi-agent pathfinding~\cite{Li2020}, flocking~\cite{Tolstaya20-Learning}, target tracking~\cite{Zhou2022}, and coverage control~\cite{Gosrich2022, Agarwal2024}.
In these examples, GNNs operate over a communication graph.
Within each GNN layer, agents update their state by aggregating information from their neighbors in this graph.
A multi-layer GNN comprises iterative information exchanges between robots with direct communication links.
GNNs have provable stability properties that enable generalization to larger team sizes~\cite{Ruiz21-GraphNeuralNetworks, Agarwal2024} and enable transferability to new tasks~\cite{Ruiz20-GraphonNeural}.
Moreover, the \emph{locality} of computations on the communication graph enables GNNs to be implemented in a distributed manner~\cite{Tolstaya20-Learning}.

However, iterative local computations introduce architectural bottlenecks that limit the performance of GNNs: over-smoothing and over-squashing. 
\emph{Over-smoothing} can lead to performance degradation for multi-layer GNNs~\cite{Alon20-BottleneckGraphNeural}. 
\emph{Over-squashing} is the observation that GNNs struggle with transmitting information across multiple hops~\cite{Topping21-Understanding}.
Additionally, distributed implementation requires that communication is modeled with sufficient accuracy during training. 
Changing communication hardware or deployment in an environment with increased noise may impact performance by altering the structure of the communication graph.

\subsection{Transformers in Multi-Agent Systems}
Several recent works leverage transformers to process visual~\cite{Chen23-TransformerBased, Chen24-TransformerBased} or temporal information~\cite{Wang23-NaviSTARSociallyAware} in the local perception model, while others focus on solving tasks centrally~\cite{Yuan21-AgentFormerAgentAware}.
In~\cite{Zhang24-Decentralized}, a transformer perceiver~\cite{Jaegle21-PerceiverGeneral} architecture is used to model the global multi-agent state and compute collaborative actions.
Wen et al.~\cite{Wen22-MultiagentReinforcement} propose the multi-agent transformer (MAT), wherein agents make decisions sequentially, reinterpreting multi-agent decision-making as a sequential problem.
In~\cite{Egorov22-ScalableMultiAgent}, transformers are implemented in a decentralized fashion by running a local model to represent observations and state of each agent, gathering all embeddings at each agent, and running a transformer locally with the gathered information. 
This strategy can be applied to settings with sporadic ad-hoc communication, as transformers can operate on variable-size inputs~\cite{Farjadnasab25-Cooperative}, similarly to GNNs.

These works demonstrate the potential use-case of transformers in multi-agent systems. 
However, they focus primarily on systems with small teams (typically fewer than 10 agents) and interpret transformers as operations over ordered sequences.
Although transformers were initially introduced for sequence modeling~\cite{Vaswani2017}, self-attention is a more general set-to-set operation~\cite{Bronstein21-GeometricDeep} that can process \emph{spatial} information~\cite{Zhao21-PointTransformer}.

This article introduces the transformer-based MAST architecture as an alternative backbone for the communication model.
Similar to GNNs, MAST can handle variably sized inputs, allowing generalizability to varying team sizes.
We also show how MAST can be deployed in decentralized settings by leveraging the communication graph.

\subsection{Positional Encodings}
The attention mechanism in transformers are designed to be \emph{permutation equivariant}, i.e., permuting the inputs, permutes the output in the same way.
However, in natural language, the positions of the tokens in a phrase is critical.
To incorporate such positional information, several encoding strategies have been proposed.
Three widely used encodings are:
\emph{(i)}~Absolute positional encodings (APE)~\cite{Vaswani2017};
\emph{(ii)}~Learnable MLP positional encodings~\cite{Zhao21-PointTransformer}; and
\emph{(iii)}~Rotary position embeddings (RoPE)~\cite{Su2024, Heo2024}.

The APE uses a series of sine and cosine functions of steadily increasing wavelength.
The learnable MLP encodings use weights at each token position that are learned jointly with the core model parameters during training.
However, this makes the encodings dependent on the positions of the encodings, i.e., they are no longer permutation invariant.
RoPE~\cite{Su2024, Heo2024} is a recently proposed way to include \emph{relative} positional information into attention.
The general idea is to multiply the inputs to attention by complex exponentials, so that the inner products used to compute attention weights depend only on the relative displacement and not on the individual absolute positions.

These strategies are designed for integral positions and do not apply directly to positions of robots in Euclidean space.
In this article, we propose a new relative position embedding strategy, inspired by RoPE, that is is specifically designed for robot positions.
We show the application of this new embeddings within the transformer architecture.



\section{Distributed and Collaborative Multi-Agent Systems}\label{sec:decentralized-robotics}

Problems within the purview of multi-robot systems span a broad gamut. 
We focus our efforts on tasks that fall under the umbrella of multi-robot navigation with a homogeneous team of robots, \(\calV=\{1,\ldots,N\}\).
We restrict our attention to the 2D case with first-order dynamics.
Denote the position of a robot~$i$ with $\bfp_i(t)\in \reals^2$ at time~$t$.
All robot positions is given by the matrix $\bfP(t)$:
\begin{equation}\label{eq:agent_position}
    \bfP(t) := \bmat{\bfp_1(t) & \bfp_2(t) & \cdots & \bfp_N(t) }^\top \in \reals^{N \times 2}.
\end{equation}
The input to the dynamical system is the velocity action $\bfu_i(t)\in \reals^2$ for robot~$i$ at time~$t$.
All robot actions are given by the matrix $\bfU(t)$:
\begin{equation}\label{eq:control}
    \bfU(t) = \bmat{\bfu_1(t) & \bfu_2(t) & \cdots & \bfu_N(t) }^\top \in \reals^{N \times 2}.
\end{equation}
Additionally, we assume that the control inputs are constrained by the maximum velocity \(u_\mathrm{max} \in \reals_+\):
\begin{equation}\label{ineq:velocity_constraint}
    \norm{\bfu_i(t)} \le u_\mathrm{max}.
\end{equation}
With $\Delta t$ as the discrete time-step, the system evolves with the following discrete dynamics:
\begin{equation}\label{eq:dynamics}
    \bfP(t + \Delta t) = \bfP(t) + \bfU(t) \Delta t.
\end{equation}

We assume that the robots are capable of self-localization, limited perception, and limited communication.
The robots can observe their local environment, which we represent as vectors $\bfo_i(t) \in \observationSpace$ in some observation space~$\observationSpace$.
The observations are local, so no robot has complete information about the environment or the multi-robot system state.
The robots can communicate with other nearby robots, but this is limited by a bandwidth constraint imposed by the communication channel~\cite{Klavins04-Communication}.
This constraint limits the scalability of centralized control frameworks.
Instead, we seek a decentralized control framework that leverages limited communication capabilities.
The framework must determine \emph{what} should be communicated and \emph{how} this information is combined by each agent to determine its action.
\saurav{List assumptions clearly.}

\begin{figure}[htbp]
    \centering
    \includegraphics[width=\columnwidth]{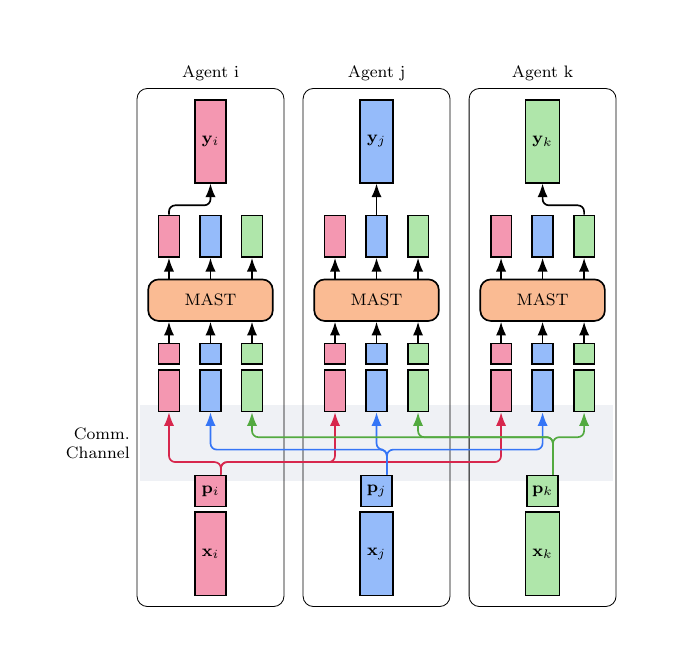}
    \caption{Idealized representation of MAST architecture, illustrating agents $i$, $j$, and $k$. A simplified communication process is depicted; phenomena such as delays and limited connectivity are neglected. The agents' positions and embeddings are communicated between agents and processed independently by each agent.}
    \label{fig:attention-decentralized}
 \end{figure}


\subsection{Learnable Perception Action Communication}

To tackle this problem, we design an architecture based on the principles of learnable perception-action-communication (LPAC) loops \cite{Hu22-Scalable, Agarwal2024}.
LPAC addresses the communication constraints by training an end-to-end model with two main components: a \emph{perception model} and an \emph{action model}. 
The \emph{perception model} learns \emph{what} should be communicated. 
It runs locally on each agent and learns an efficient latent representation $\bfx_i \in \reals^d$ of the raw observations $\bfo_i$.
These embeddings $\bfx_i$ are asynchronously communicated between agents.
An \emph{action model} learns \emph{how} to combine multiple received embeddings.
It runs locally, processing a set of embeddings into an action.

Previous works~\cite{Agarwal2024, Khan2021, Gosrich2022, Hu22-Scalable} use graph neural networks as the backbone for the action model.
This work introduces a transformer-based backbone: the multi-agent spatial transformer (MAST).
MAST is designed to be capable of centralized training, decentralized execution, and generalizing to larger teams of agents.
\fgref{fig:architecture-abstract} illustrates the proposed architecture from the perspective of the $k$th agent, and each component is described below.

The purpose of the perception model is to find $\bfx_k(t) \in \reals^{d}$, a latent vector representation of a robot's observations $\bfo_k(t)$.
The embeddings are computed locally by each robot from the observation vector:
\begin{equation}\label{eq:local_model}
    \bfx_k(t) = \perceptionModel(\bfo_k(t), \calW_P)
\end{equation}
where $\calW_P$ is a set of learnable parameters.
The parameters $\calW_P$ are shared between robots.
Each agent computes their own $\bfx_k(t)$ using a shared parameter set $\calW_P$ and their local observations $\bfo_k(t)$.
The local model is task-specific because the observation space $\observationSpace$ may vary between tasks.
For example, if observations are numerical, we could use a multi-layer perceptron, but if observations are images, we may use a convolutional neural network.
Regardless of the architecture, the perception model outputs $\bfx_k(t)$, a latent representation of the observations.

The embeddings $\bfx_k$ are exchanged between agents following some asynchronous communication process.
Our proposed method is agnostic to the specifics of this process, but we assume that the $k$th robot receives a copy of the embeddings, represented by a matrix $\bfX_k \in \reals^{N \times d}$. 
The rows of $\bfX_k$ are the latent vectors from different robots, which we denote as 
\begin{equation}
    \bfX_{ki}(t) = \bfx_i(t - \tau_{ki})
\end{equation}
which may be received with a delay of $\tau_{ki}$ seconds.
Similarly, each agent receives corresponding positions, represented by $\bfP_k \in \reals^{N \times 2}$. 

At a high level, MAST is a transformer model that uses self-attention to process the received embeddings $\bfX_k$ localized at positions $\bfP_k$. 
MAST runs independently on each agent, producing embeddings $\bfY_k(t) \in \reals^{N \times d}$ from the received information. 
Therefore, we denote MAST as a function
\begin{equation}\label{eq:mast}
    \bfY_k(t) = \operatorname{MAST}(\bfP_{k}(t), \bfX_{k}(t), \calW_\mathrm{MAST})
\end{equation}
where $\calW_\mathrm{MAST}$ is a set of parameters shared between agents.
MAST maps embeddings $\bfX_k$ localized at $\bfP_k$ to a new set of embeddings $\bfY_k$.
The rows of $\bfX_k$, $\bfP_k$, and $\bfY_k$ have a one-to-one correspondence so that $\bfY_{ki}$ is a new representation of $\bfX_{ki}, \bfP_{ki}$.
To compute the next action, each robot projects their own representation $\bfY_{kk} \in \reals^d$ to an action $\bfu_{k}$ using an MLP:
\begin{equation}\label{eq:action_model}
    \bfu_k(t) = \mathrm{MLP}(\bfY_{kk}(t), \calW_U)
\end{equation}
where $\calW_U$ is a set of parameters shared by all robots.

\section{Multi-Agent Spatial Transformer}\label{sec:mast}

In this section, we describe the multi-agent spatial transformer (MAST).
The proposed architecture is capable of decentralized processing of data coming from multiple agents at disparate locations.
This section is organized as follows. 
\scref{subsec:self-attention} describes the basic transformer architecture in the context of multi-agent systems, comprising multiple self-attention layers.
\scref{subsec:rope} describes rotary positional encodings, which allow the transformer to integrate spatial information efficiently and \emph{shift-equivariantly}. 
\scref{subsec:windowed-attention} introduces a spatial attention window that \emph{localizes} processing spatially, improving transformer scalability.
Finally, \scref{subsec:decentralized_inference,subsec:centralized_training} describe how MAST is capable of decentralized execution and efficient centralized training, respectively.

\subsection{Transformer}\label{subsec:self-attention}

Self-attention \cite{Vaswani2017} is a mechanism that processes a set of vectors by considering the relationships between all elements in the set. 
Typically, the inputs to self-attention are vectors $\bfx_i \in \reals^d$ for $i \in \{1,\dots,N\}$ and the outputs are vectors $\bfy_i \in \reals^d$ for $i \in \{1,\dots,N\}$. 
At a high level, self-attention is computed as a weighted sum of the inputs $\bfx_i$. 
The attention weights are computed parametrically from pairs of input vectors. A common parametrization is the inner product between projections:
\begin{equation}\label{eq:sdpa_product}
    a_{ij} = \langle \bfQ \bfx_{i}, \bfK \bfx_{j} \rangle
\end{equation}
where $\bfQ, \bfK \in \reals^{d_a \times d}$ are projection matrices to a lower dimensional space $d_a$.
The self-attention is then defined as:
\begin{equation}
    \bfy_{i} = \frac{\sum_{j=1}^{N} \exp(a_{ij}) \bfV \bfx_j }{\sum_{j=1}^{N} \exp(a_{ij})} \label{eq:sdpa_sum}
\end{equation}
where $\bfV \in \reals^{d_a \times d}$ is a projection matrix, and the attention weights are softmax-normalized.

Multi-head self-attention (MHSA) is a common extension of self-attention. In MHSA, multiple self-attention operations are computed in parallel on the GPU; their outputs are concatenated and projected again. Each parallel self-attention operation is known as a \emph{head}. Each head has its own independent set of learnable parameters. Let $\bfy_i^h$ be the output of the $h$th attention head. Then, the output of MHSA is given by:
\begin{equation}\label{eq:mhsa}
    \bfy_i = \bfO \left( \bfy_i^1 \| \bfy_i^2 \| \ldots \|  \bfy_i^H \right)
\end{equation}
where $\bfO \in \reals^{d \times d_a H}$ is a learnable matrix, and the output of each head is concatenated and then combined using a linear transformation.

The MAST architecture has $L$ layers comprising MHSA and an MLP.
Each MHSA layer uses $H$ attention heads with head dimension $d_a = d / H$ and $d$-dimensional inputs and outputs.
This is followed by a single-layer MLP with hidden dimension $2d$, input dimension $d$, and output dimension $d$. 
MAST uses standard pre-normalization \cite{Xiong20-LayerNormalization} residual connections, applied to embeddings after each attention and MLP layer. 
Leaky ReLU activations are used throughout the architecture.

\subsection{Rotary Positional Encoding}\label{subsec:rope}
\damian{Focus more on how we are implementing and the specific requirements for DCMRS.}
Positional encodings are a way to include spatial information about the inputs during self-attention.
RoPE \cite{Su2024, Heo2024} is a recently proposed way to include \emph{relative} positional information into attention.
Using RoPE, the attention weights $a_{ij}$ depend only on the relative displacement, $\bfp_i - \bfp_j$, and not on the individual absolute positions.
Here, we extend RoPE to continuous positions in 2D, even though it was initially proposed for discrete positions in language~\cite{Su2024} and vision~\cite{Heo2024} tasks.

Mathematically, we will describe RoPE with complex-valued projections to $\mathbb{C}^{d_a/2}$.
In practice, we can implement RoPE with real-values projections in $\mathbb{R}^{d_a}$, as demonstrated in \cite{Su2024}, by interpreting alternating elements of the keys and queries as the real and imaginary parts of a complex number.
Before attention, RoPE multiplies $\bfx_i$ by a vector of complex-exponential functions of $\bfp_i$.
Denote this complex-exponential vector as $\bfr(\bfp_i) : \reals^2 \mapsto \mathbb{C}^{d_a/2}$.

The components of $\bfr(\bfp_i)$ are complex-exponentials with frequencies $\omega_k \in \reals_+$:
\begin{equation}\label{eq:rope_complex_exponential}
    \bfr(\bfp_i) = \begin{bmatrix}
        \vdots           \\
        r_{2k}(\bfp_i)   \\
        r_{2k+1}(\bfp_i) \\
        \vdots
    \end{bmatrix} = \begin{bmatrix}
        \vdots                 \\
        \exp(j \omega_k p_i^x) \\
        \exp(j \omega_k p_i^y) \\
        \vdots
    \end{bmatrix}
\end{equation}
\damian{Make sure that you don't mention APE.}
where $r_{n}(\bfp_i)$ is the $n$th element of $\bfr(\bfp_i)$, $j = \sqrt{-1}$, and $\omega_k \in \reals_+$ are the same angular frequencies as in APE; see \eqref{eq:frequencies_geometric} and \eqref{eq:frequencies_linear}.
Therefore, RoPE computes attention weights differently from Equation~\eqref{eq:sdpa_product}:
\begin{equation}\label{eq:rope1}
    a_{ij} = \Re\left[ \langle \bfr(\bfp_i) \odot \bfQ \bfx_i, \bfr(\bfp_j) \odot \bfK \bfx_j \right]
\end{equation}
where $\odot$ denotes element-wise multiplication, and the linear projections $\bfQ, \bfK \in \mathbb{C}^{(d_a/2) \times d}$ are complex matrices. 
The attention weights are the \emph{real-part} of the complex dot product. 

\subsubsection{Shift-equivariance of RoPE}
\damian{Explain why this is important.}
We can establish the shift-equivariance of RoPE with some algebraic manipulation. In other words, we can show that the attention weight $a_{ij}$ in~\eqref{eq:rope1} for RoPE depends only on the relative position $\bfp_{ij}=\bfp_i - \bfp_j$ and not on either of the absolute positions, i.e., \eqref{eq:rope1} can be equivalently written as:
\begin{equation}\label{eq:rope2}
    a_{ij} = \Re\left[ \langle \bfr(\bfp_{ij}) \odot \bfQ \bfx_i, \bfK \bfx_j \rangle \right]
\end{equation}
To establish this equivalence, consider the definition of the complex dot product. The dot product is the sum of the element-wise product between the left side and the conjugate of the right side. Since $\bfr(\bfp_j)$ is multiplied element-wise, we can use associativity and commutativity of the element-wise product to move it from the left side to the right side of the inner product. When doing so, we need to conjugate it, but since it is a complex exponential, the conjugate of $\bfr(\bfp_j)$ is $\bfr(-\bfp_j)$.

\subsubsection{Choice of Frequencies}\label{subsec:positional_encoding_frequencies}
\damian{Make discussion about previous choice of encoding shorter.}
The frequencies $\omega_k$ are commonly chosen \textit{a priori}. The original transformer architecture \cite{Vaswani2017} uses a geometric series,
\begin{equation}
    \omega_k = 10000^{-k / K}
\end{equation}
where $K$ is the total number of frequencies.
The smallest frequency $\omega_K = 10000^{-1}$ is typically known as the base frequency, and the corresponding wavelength is $\baseWavelength = 2\pi\omega_k = 2\pi \times 10000 \approx 62,831$, which we call the \emph{base wavelength}.
Note that $\baseWavelength$ was explicitly chosen for machine translation tasks and experiments where the input sequence length was approximately $25000$ tokens~\cite{Vaswani2017}.
In this work, we study applications to multi-robot systems, so we make a different choice for base wavelength.
We can rewrite the above geometric series in terms of $\baseWavelength$:
\begin{equation}\label{eq:frequencies_geometric}
    \omega_k = 2\pi \baseWavelength^{-k / K}
\end{equation}
In our experiments, we set $\baseWavelength$ proportional to the width of the environment.

\begin{figure}
    \centering
    \includegraphics[width=\linewidth]{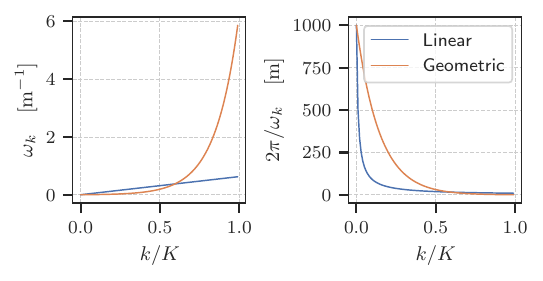}
    \caption{Plot of the frequencies and the corresponding period for the two different frequency sets: the linear frequency set and the geometric frequency set described by Equations \eqref{eq:frequencies_linear} and \eqref{eq:frequencies_geometric}, respectively.
    A positional encoding that uses the geometric frequency set has more sinusoids with a larger period. Relative to the linear frequencies, it will have a higher resolution to distinguish between faraway points than between nearby points.
    \damian{Consider removing and including in text.}
    }
    \label{fig:frequencies}
\end{figure}

While the original transformer architecture uses a geometric series to define the frequencies \cite{Vaswani2017}, we experiment with an alternative set of frequencies. We hypothesize that the optimal way to choose frequencies may vary between tasks, just like the optimal base frequency is task-dependent.
We experiment with linearly spaced frequencies as an alternative to the set of geometrically spaced frequencies.
\begin{equation}\label{eq:frequencies_linear}
    \omega_k = \frac{2\pi k}{\baseWavelength}.
\end{equation}
Analogous to the geometric frequency set in~\eqref{eq:frequencies_geometric}, the above expression~\eqref{eq:frequencies_linear} specifies a linear frequency set, where the frequencies are evenly spaced between $0$ and $\baseWavelength$.
As with the geometric frequency set, we use $\baseWavelength$ equal to the width of the training environment.
\fgref{fig:frequencies} shows how the frequencies and wavelengths generated from \eqref{eq:frequencies_geometric} and \eqref{eq:frequencies_linear} compare.

\damian{Try making it much shorter. 1-3 sentences if possible.}
In our experiments, we set $\baseWavelength$ equal to the width of the training environment, denoted by $W$.
We argue that setting $\baseWavelength = W$ maximizes the information content of the positional encodings. Without loss of generality, assume that the positions, $\bfp_i \in [0, W]^2$, are non-negative and bounded during training.
Consider absolute positional encoding (APE) with linearly spaced frequencies.
These encodings are periodic with period $\baseWavelength$.
If $\baseWavelength < W$, then the positional encodings will \emph{repeat} or \emph{alias}, i.e., two distant positions may share the same encoding.
Aliasing is undesirable, as it causes the model to confuse different positions.
To avoid aliasing, we want $\baseWavelength \geq W$.

However, if we choose $\baseWavelength > W$, the encodings will include positions with $\bfp_i \notin [0, W]^2$, which the model never sees during training.
This can be detrimental to the effective resolution of the model and its ability to generalize, especially to larger environments.
Therefore, we use $\baseWavelength = W$ as a tradeoff; it avoids aliasing without introducing untrained positional encodings.

This argument has two caveats.
First, if we want the model to generalize to larger environments and use $\baseWavelength > W$, aliasing will still occur at test time. We mitigate this by using an attention window: as long as the model’s receptive field is smaller than $W$, aliasing within that window does not happen.
Second, the argument above assumes linearly spaced frequencies. The encoding is not strictly periodic on $\baseWavelength$ for geometrically spaced frequencies, so the aliasing behavior changes.

\subsection{Windowed Attention}\label{subsec:windowed-attention}

This work introduces radially windowed self-attention based on the distance between agents $\lVert \bfp_i - \bfp_j \rVert$.
\damian{As we show in Section~\ref{subsec:dan_gridserach}, radially windowed attention improves the generalizability of a model to larger teams than seen in training.}
Additionally, such windowing could improve scalability for large agent teams by eliminating the need to compute inner products between faraway agents.
In \eqref{eq:sdpa_sum}, the attention mechanism attends \emph{globally} to all other agents.
Global attention patterns may impede generalization to larger teams of agents after training with smaller teams~\cite{Owerko23-Transferability}.
Additionally, since the computational complexity of global attention is quadratic in the number of agents, this may make it unscalable computationally.
Instead, we propose to window attention radially, based on the Euclidean distance $\lVert \bfp_i - \bfp_j \rVert_2$ between agents.
Windowing sparsifies the attention matrix and reduces the number of inner products we need to compute.

To implement windowing, we introduce an attention mask $M_{ij} \in \{0, 1\}$, which indicates whether the attention weight between $i$ and $j$ should be computed.
The self-attention mechanism from Equation~\eqref{eq:sdpa_sum} is redefined as:
\begin{equation}\label{eq:sdpa_sum_mask}
    \bfy_{i} = \frac{\sum_{j=1}^{N} M_{ij} \exp(a_{ij}) \bfV \bfx_j }{\sum_{j=1}^{N} \exp(a_{ij})}
\end{equation}
where the mask is given by
\begin{equation}\label{eq:attention_mask}
    M_{ij} = \mathds{1}( \lVert \bfp_i - \bfp_j \rVert_2 < R_{\text{att}} ).
\end{equation}
$R_{\text{att}}$ is the radius of the attention window.
Equation \eqref{eq:sdpa_sum_mask} uses the mask to ignore the contributions of agents that are too far away.
In \scref{sec:dan}, we report experimental results showing how this windowing impacts size generalizability.

\subsection{Decentralized Execution}\label{subsec:decentralized_inference}

During decentralized execution, communication bandwidth, latency, and range limit the available information to each agent.
Above, we have implicitly assumed that each agent has access to the embeddings and positions of all other agents in the system.
Instead, we make two \red{assumptions} about decentralized execution: that each agent receives messages from a subset of other agents, and that the messages arrive with a delay.
The transformer architecture naturally accommodates this, since the parametrization is independent of the number of rows in the input matrices.

Consider the $k$th agent and let $\calV_k(t) \subseteq \calV$ be the set of all agents from which agent $k$ has received information between time $0$ and $t$.
Let $\calI_k: \{1,\dots,N_k\} \to \calV_k(t)$ be an arbitrary ordering of the elements of $\calV_k$.
Therefore, we denote the embeddings available to the $k$th robot as a matrix $\bfX_k(t) \in \reals^{N_k \times d}$ with rows $\bfX_{ki}(t) \in \reals^d$:
\begin{equation}\label{eq:gathered_embeddings}
    \bfX_{ki}(t) = \bfx_{\calI_k(i)}(t-\tau_{ki}(t))
\end{equation}
for $i \in \{1,\dots,N_k\}$ where $\tau_{ki}(t)$ is the delay at which information from agent $\calI_k(i)$ arrives at agent $k$.
Similarly, the matrix $\bfP_k \in \reals^{N_k \times 2}$ contains the positions of other agents from the perspective of the $k$th agent:
\begin{equation}\label{eq:gathered_positions}
    \bfP_{ki}(t) = \bfp_{\calI_k(i)}(t-\tau_{ki}(t))
\end{equation}
for $i \in \{1,\dots,N_k\}$
MAST operates only on the locally available information from this reduced subset of agents.
As in \eqref{eq:mast}, the inputs are $\bfX_k(t)$ and $\bfP_k(t)$, and the output is $\bfY_k(t) \in \reals^{N_k \times d}$.
Assuming without loss of generality that the rows are ordered so that the local agent information is first, we can rewrite~\eqref{eq:action_model}:
\begin{equation}\label{eq:action_model_decentralized}
    \bfu_k(t) = \mathrm{MLP}(\bfY_{k1}(t), \calW_U),
\end{equation}
where $\bfY_{k1}(t)$ represents the $k$th agent.

Although MAST is agnostic to the communication topology, we assume that the robots communicate over some multi-hop communication network. 
Robots can be readily equipped with multi-hop communication capability using off-the-shelf hardware \cite{Mox24-Opportunistic}.
In our experiments, we assume the existence of some dynamic communication graph $\calG(t) = (\calV, \calE(t))$. 
Nodes in the graph $\calV = \{1, \dots, N\}$ are agent indices with edges $(i,j) \in \calE(t)$ if and only if agent $i$ can directly communicate with agent $j$ at time $t$.
The topology of $\calG(t)$ differs between our experiments to highlight the robustness of our architecture. Section~\ref{sec:dan} uses a k-nearest neighbor graph, while Section~\ref{sec:coverage_control} uses a distance threshold graph.
Additionally, we model the latency of multi-hop communication.
We assume that information propagates across $\calG(t)$ with a time delay of $\tau$ seconds per hop.
We simulate this delayed propagation using Algorithm~\ref{alg:communication}, where agents buffer and relay timestamped messages.
Agents use timestamps to keep only the most recent embedding per agent.


\begin{algorithm}
    \caption{Simulation of multi-hop communication.}
    \label{alg:communication}
    \begin{algorithmic}
        \State Let $t \gets 0$ be the start time.
        \State Let $\bfX_{ki}(t), \bfP_{ki}(t)$ be the most recent embedding and positions originating from agent $i$ that agent $k$ has received on or before time $t$.
        \State Let $t_{ki}(t) \gets -1$ be timestamp of the most recent message originating from agent $i$ that agent $k$ has before $t$.
        \While{true}
        \For{$k \in \calV$}
        \Comment{Update information about self.}
        \State $\bfX_{kk}(t) \gets \bfx_k(t)$
        \State $\bfP_{kk}(t) \gets \bfp_k(t)$
        \State $t_{kk}(t) \gets t$
        \EndFor
        \State $t \gets t + \tau$
        \Comment{Communication process takes $\tau$ seconds.}
        \For{$k \in \calV$}
        \For{$j \in \calN(k)$}
        \Comment{Neighbors in comm. graph.}
        \State \Call{Receive}{$k$, $j$}
        \EndFor
        \EndFor
        \For{$k \in \calV$}
        \State $\calX_k \gets \{\bfX_{ki}(t) \mid i \in \calV, t_{ni} \ge 0\}$
        \State $\calP_k \gets \{\bfP_{ki}(t) \mid i \in \calV, t_{ni} \ge 0\}$
        \EndFor
        \State \textbf{yield} $\calX_k, \calP_k$
        \EndWhile
        \Function{Receive}{$k$, $j$}
        \For{$i \in \calV$}
        \If{$t_{ki}(t) < t_{ji}(t)$}
        \State $\bfX_{ki}(t) \gets \bfX_{ji}(t-\tau)$
        \State $\bfP_{ki}(t) \gets \bfP_{ji}(t-\tau)$
        \State $t_{ki}(t) \gets t_{ji}(t-\tau)$
        \EndIf
        \EndFor
        \EndFunction
    \end{algorithmic}
\end{algorithm}

\subsection{Centralized Training}\label{subsec:centralized_training}

\begin{figure*}
    \centering
    \includegraphics[width=0.8\linewidth]{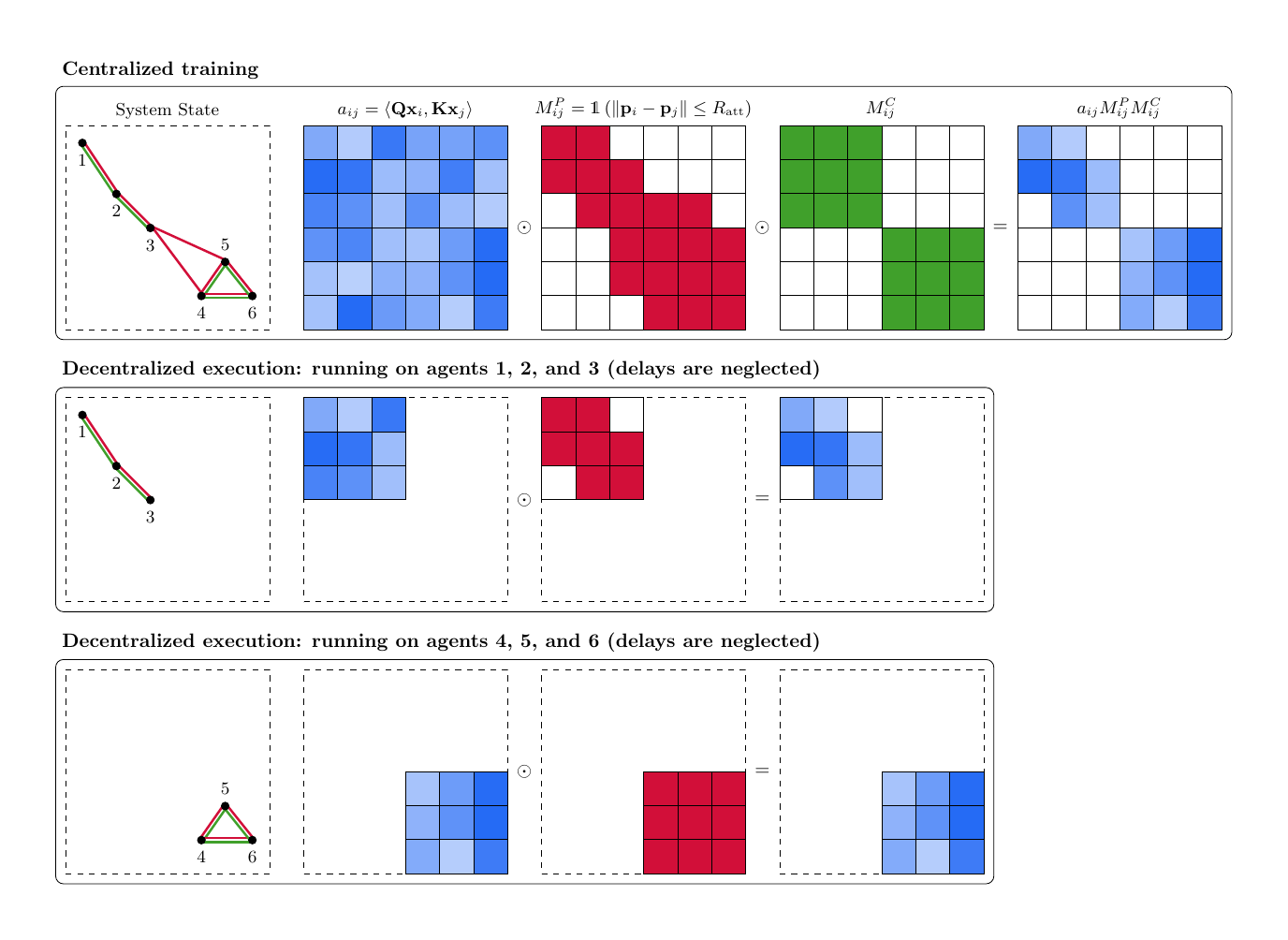}
    \caption{
    Illustration of the difference in attention during centralized training and decentralized inference.
    The communication graph is shown in green and has two connected components: $\{1,2,3\}$ and $\{4,5,6\}$.
    The first row shows centralized training, where the component mask $M_{ij}^C$ prevents agents in different connected components from attending to each other.
    The second and third rows show the attention weights and masks during decentralized execution for different connected components.
    Agents can only access information within their own connected component and run MAST using that subset of embeddings and positions.
    }
    \label{fig:attention-mask}
\end{figure*}

During decentralized execution, MAST is executed locally by each agent, which requires executing MAST $N$ times.
We designed an efficient centralized training methodology that approximates the impact of the communication topology, while only running once for \emph{all} agents simultaneously.
Our primary optimization during training is to ignore the effects of communication delays.
When communication is instantaneous, the received embeddings and positions are the same across all agents within each connected component of $\calG$. 
Therefore, we only need to run MAST once per connected component, instead of once per agent.

We optimize this further by parallelizing computations of attention over several connected components.
To accomplish this, we introduce a \emph{component mask} with elements $M_{ij}^C \in \{0,1\}$, which are non-zero whenever $i$ and $j$ are in the same connected component:
\begin{equation}
    M_{ij}^C = \begin{cases}
        1 & \text{if there exists a path from } j \to i \text{ in } \calG \\
        0 & \text{otherwise}.
    \end{cases}
\end{equation}
By using the component mask in~\eqref{eq:sdpa_sum_mask}, attention is only computed between agents in the same connected component.
With component masking, running MAST on all agents is equivalent to running MAST once per connected component.
Note that we perform both component masking and attention windowing, which was introduced in Section~\ref{subsec:windowed-attention}.
Let $M_{ij}^P$ be the mask from~\eqref{eq:attention_mask} so that
\begin{equation}
    M_{ij}^P = \mathds{1}( \lVert \bfp_{i} - \bfp_{j} \rVert_2 < R_{\text{att}} ).     \label{eq:positional_mask}
\end{equation}
Therefore, during centralized training, we use the bitwise AND between the positional and component masks:
\begin{equation}
    M_{ij} = M_{ij}^P M_{ij}^C, \label{eq:attention_mask_mast}
\end{equation}
which is used in \eqref{eq:sdpa_sum_mask}.
Incorporating the component mask allows us to reduce the distributional shift between centralized training and decentralized execution.

Designing centralized training to closely approximate decentralized execution is critical to maintain generalizability.
Figure~\ref{fig:attention-mask} illustrates how component masking accomplishes this.
In the first row of the figure, the component mask (shown in green) ensures that attention is computed independently for each connected component.
After masking, the attention coefficients $a_{ij}M_{ij}^PM_{ij}^C$ are zero whenever agents $i$ and $j$ are in different connected components on $\calG$.
Agents must be in the same connected component and within the attention radius for attention to be non-zero between them.
During decentralized execution, the component mask $M_{ij}^C$ is implicit. 
The first column of Figure~\ref{fig:attention-mask} uses green edges to depict the communication graph, $\calG$. 
The graph has two connected components: agents $\{1,2,3\}$ and $\{4,5,6\}$. 
Agents receive information only from others in the same connected component.
In summary, when delays are negligible, component masking is equivalent to executing MAST locally by each agent.

\damian{Wrap up the section instead of ending abruptly.}

\section{Decentralized Assignment and Navigation}\label{sec:dan}
\newcommand{\goalRadius}{R_{\mathrm{g}}}

This section presents the application of the proposed MAST archnitecture to the decentralized assignment and navigation (DAN) problem.
The problem is defined on a system of $N$ robots and $N$ goal positions, and the task is to simultaneously assign robots to goal positions and navigate to their respective goals.
The locality and decentralization of the system makes the problem challenging in the following ways:
\emph{(i)}~Each robot has knowledge of goal positions only within a certain range, and
\emph{(ii)}~Each robot takes navigation actions independently, based on its local perception and the information received from other nearby agents.

The results presented in this section highlights the following aspects of the MAST architecture incorporated into the LPAC policy:
\emph{(i)}~The learned policy performs better than decentralized Hungarian algorithm and achieves a success rate of about $85$\%.


\subsection{Problem Formulation}\label{subsec:dan_problem_formulation}

The problem is described mathematically as follows.
Consider positions $\bfP(t)$, control inputs $\bfU(t)$, and system dynamics from Equations \eqref{eq:agent_position}, \eqref{eq:control}, and \eqref{eq:dynamics} in Section~\ref{sec:decentralized-robotics}.
In addition, let
\begin{equation}\label{eq:goal_position}
    \bfG = \bmat{\bfg_1 & \bfg_2 & \cdots & \bfg_N }^T \in \reals^{N \times 2}
\end{equation}
represent the goal positions.
To quantify the quality of a solution, we introduce \emph{success rate} in Equation~\eqref{eq:success_rate}. 
It is defined as the proportion of goals for which an agent is within a threshold distance \(\goalRadius\).
\begin{equation}\label{eq:success_rate}
    \mathrm{SR}(t) = \frac{1}{N} \sum_{i=1}^N \mathds{1}(\min_j \norm{\bfg_i-\bfp_j}_2 < \goalRadius).
\end{equation}
We can calculate the success rate by finding the closest agent to each goal and checking if the distance is less than \(\goalRadius\).

\damian{
\begin{figure}
    \centering
    \begin{tikzpicture}
    \definecolor{mBlue}{HTML}{1F77B4}
    \definecolor{mDarkRed}{HTML}{a2282f}
    \definecolor{mSteelGray}{HTML}{CACACA}
    \footnotesize
    \tikzstyle{Arc} = [thick,-{Latex[length=2mm,width=1.8mm]}]
    \tikzstyle{robot}=[circle,draw=mBlue,thick,fill=white,inner sep=0.4mm, outer sep=0, text=mBlue]
    \tikzstyle{target}=[circle,thick,draw=mDarkRed,fill=white,inner sep=0.4mm, outer sep=0, text=mDarkRed]
    \tikzstyle{arrow}=[black, -Stealth, thick]
    \begin{axis}[xmin=-0.5,xmax=5.5,ymin=-0.5,ymax=1.5, xtick distance=1, ytick distance = 1, height=2in, width=3.5in, xlabel=$x$, ylabel=$y$]
        \node (r1) at (2,1) [robot]{$1$};
        \node (r2) at (0,0) [robot]{$2$};
        \node (t1) at (3,1) [target]{$1$};
        \node (t2) at (5,0) [target]{$2$};

        \draw[arrow] (r1) -- (t1) node[pos=0.5,above]{$1$};
        \draw[arrow] (r2) -- (t2) node[pos=0.5,above]{$5$};
        \draw[arrow, dashed] (r1) -- (t2) node[pos=0.5,sloped, above]{$\sqrt{10}$};
        \draw[arrow, dashed] (r2) -- (t1) node[pos=0.5,sloped, above]{$\sqrt{10}$};
    \end{axis}

\end{tikzpicture}
    \caption{Difference between sum of distance-squared and sum of distance assignments. Agents in blue are assigned to goals in red. The lowest total distance traveled assignment is shown using solid lines, while the distance squared assignment uses dashed lines. 
    The distance-squared assignment is more equitable.
    }
    \label{fig:assignment_comparison}
\end{figure}
}
\subsubsection{Centralized Policies}\label{subsec:dan_centralized_policies}

\damian{No need to go into this much detail about sum of distances vs sum of distance-squared assignment.}
A common way to address the assignment and navigation problem is to find a matching between agents and goals and then navigate to the assigned goals using a control policy. Agents move directly toward the assigned goal at maximum velocity.
Prior works use a matching that minimizes the sum of distance-squared traveled \cite{Turpin2014, Panagou2014, Goarin2024}. Minimizing distance-squared is more popular than minimizing total distance traveled.
\damian{Figure~\ref{fig:assignment_comparison} illustrates the difference between them.}
Distance-squared assignment trades off higher total distance for reductions in per-agent distance variance. The trade-off is analogous to the trade-off between L2 and L1 regularized regression \cite{Zou05-RegularizationVariable}. Prior theoretical work also provides closed-form collision-free solutions for \emph{centralized} assignment and navigation with a distance-square assignment \cite{Turpin2014}.
Therefore, throughout this work, we focus on \emph{distance-squared assignment} and remain consistent with the literature.

We can formulate a linear sum assignment problem (LSAP)~\cite{Kuhn1955} with distance-squared as the cost matrix to find an assignment that minimizes the sum of distances-squared.
Let \(\bfD(t) \in \reals^{N \times N}\) be a matrix with entries \(D_{ij} = || \bfp_i(t) - \bfg_j ||_2^2\) that are equal to the distance-squared between the $i$th agent and $j$th target.
Then, the solution to the related LSAP problem is a permutation matrix,
\begin{equation}\label{eq:assignment_lsap}
    \bfS_\mathrm{LSAP}(t) = \argmin_{\bfS \in \calS_N} \mathbf{1}^T(\bfS \odot \bfD(t))\mathbf{1}
\end{equation}
where \(\calS_N\) is the set of \(N \times N\) permutation matrices and \(\odot\) is the Hadamard product. This assignment induces a max-velocity policy where each agent travels at maximum velocity with velocities parallel to $\bfS_\mathrm{LSAP}(t)\bfG - \bfP(t)$. We report the performance of the \emph{LSAP policy} to provide a reference upper bound performance in terms of $\mathrm{SR}(t)$.

\subsubsection{Decentralized Policies}\label{subsec:dan_decentralized_policies}

We focus on \emph{decentralized} assignment and navigation, where each agent only has access to local information about the environment.
We assume that each agent can observe the $K=3$ closest agents and goal positions.
Note that this is more restrictive than considered in previous works \cite{Turpin2014, Panagou2014, Ismail2017, Goarin2024}, which provide the positions of \emph{all} goals to each agent.
Many previous works focus on this less restrictive case, and \cite{Goarin2024} provides an excellent overview of such decentralized approaches.
In \cite{Goarin2024}, the authors propose to use a graph neural network to find an assignment between agents and goals.
We highlight several other approaches that were used as benchmarks in their work.
D-CAPT is a decentralized algorithm that uses pairwise swapping of goal assignments to find collision-free trajectories \cite{Turpin2014}.
Local Hungarian (LH) is a decentralized algorithm that formulates Lyapunov functions, which are used as cost functions in a local LSAP problem.
The decentralized Hungarian-based algorithm (DHBA) \cite{Ismail2017} solves the LSAP problem for each agent using information from a k-hop neighborhood.
Note that, in terms of optimality-gap, the DHBA is the best performing decentralized algorithm in \cite{Goarin2024} if given information from a sufficiently large neighborhood.

Considering the high performance of DHBA in \cite{Goarin2024}, we use it as a benchmark for our proposed approach.
In our implementation of DHBA, we assume that each agent can observe some subset of all the agents and goals.
For each agent~$i$, we run a breadth-first search over the communication graph~$\calG$ with a depth of~$k$.
Hence, we collect sets of observable agents and goals in the $k$-hop neighborhood of each agent and solve the distance-squared LSAP problem using this information.
Note that, unlike Algorithm~\ref{alg:communication} used in experiments for MAST, DHBA does not account for time delays.
We use Numba \cite{Lam15-NumbaLLVMbasedPython} to parallelize this computation.
This implementation generalizes the original DHBA to our limited information setting.
At each time step, we run this algorithm to find a matching and then set the control input of each agent to be the maximum velocity that moves them towards their assigned goal.
We refer to this family of policies as \emph{DHBA-k}.

\subsection{Simulation Environment}\label{subsec:simulation_environment}

\begin{figure*}
    \centering
    \includegraphics[width=\figwidthDouble\linewidth]{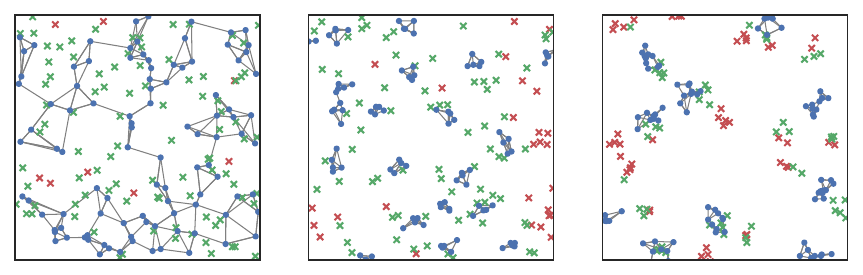}
    \caption{Example initializations of agents and goals with $\rho_\mathrm{agent} = 1, 5, 10$ agents per cluster and $\rho_\mathrm{goal} = 1, 1, 5$ goals per cluster, from left to right.
        Blue circles represent agents, green \green{$\bm{\times}$}'s represent goals observed by at least one agent, and red \red{$\bm{\times}$}'s represent goals unobserved by any agent.
        Lines represent the communication graph $\calG$.}
    \label{fig:motion-planning-initialization}
\end{figure*}

We rely on numerical simulations to train and evaluate the proposed transformer architecture.
We also use the same simulations to compare the performance against the LSAP and DHBA-k policies.
The dynamics of the multi-agent system are discretized with a time step of $\Delta t = \SIs{1.0}$.
For training, we simulate the system for $T=\SIs{200}$.
Robots have a radius $R=\SIm{2.5}$  with a maximum velocity $u_\mathrm{max} = \SIvel{5.0}$.
An agent is considered to have successfully reached a goal if it stays within a threshold radius of $\goalRadius=\SIm{5.0}$.
The communication graph $\calG(t)$ is dynamic and depends on the agent positions.
In this task, we use a directed 3-nearest neighbor graph. That is, each agent can directly receive information from its three closest neighbors.

Simulation environments are initialized with $N=100$ agents and $N=100$ goals in a $\SIkm{1} \times\SIkm{1}$ square region.
We ensure that no two agents and no two goals are within a distance of $R_\mathrm{min} = \SIm{5.0}$ of each other.
We arrange the agents into clusters of size $\rho_\mathrm{agent}$, chosen randomly from $\rho_\mathrm{agent} \in \{1, 5, 10\}$.
Similarly, we arrange the goals into clusters of size $\rho_\mathrm{goal} \in \{1, 5, 10\}$.
Note that the cluster sizes can be different for agents and goals.
Figure \ref{fig:motion-planning-initialization} shows three examples of such an initialization with varying cluster sizes.
We sample $N / \rho$ cluster centers from a uniform distribution over the square simulation region.
For each cluster center, we sample $\rho$ points from a uniform distribution over a disk of radius $r_\mathrm{cluster} = 5 R_\mathrm{min} \sqrt{\rho}$.
To enforce the minimum separation, we resample pairs of cluster centers within $2 r_\mathrm{cluster}$ until no conflicts occur.
Similarly, for any pair of points within $R_\mathrm{min}$ of each other, we resample their positions until there are no conflicts.

\subsection{Architecture}

Section~\ref{sec:mast} fully describes MAST and the readout MLP, but the perception model is task-specific. For DAN, the perception model is an MLP with parameters shared across all agents. 
The inputs are the different observations made by each agent.
Specifically, the input to the perception model of the $n$th agent is a vector $\bfo_n(t) \in \reals^{14}$ containing:
\emph{(i)}~the previous velocity of the agent,
\emph{(ii)}~the \emph{relative} positions of the 3 closest agents, and
\emph{(iii)}~the \emph{relative} positions of the 3 closest goals.
The perception MLP has one hidden layer with dimension $2d$ and leaky ReLU activations.
Recall that the output of the perception model is an embedding $\bfx_k(t) \in \reals^d$ for each agent. 
The dimensionality of the embeddings, $d$, is a hyperparameter.
Overall, this is a simple perception architecture; our focus is to demonstrate the applicability and explore the design space of transformer models rather than to optimize for a specific application.
Section~\ref{sec:coverage_control} demonstrates an application where a more complex, convolutional model is used instead.

In our numerical experiments, we compare several positional encoding schemes.
Our proposed approach is a continuous extension of RoPE, as proposed in Section~\ref{subsec:rope}.
However, we compare RoPE against two other positional encoding schemes: MLP positional encoding (MLP-PE) and absolute positional encoding (APE).
Detailed descriptions of MLP-PE and APE are in Appendix~\ref{subsec:mlp-pe} and Appendix~\ref{subsec:ape}, respectively.
Additionally, we consider both linear and geometric frequency sets for RoPE and APE, as described by Section~\ref{subsec:positional_encoding_frequencies}.
Table~\ref{tab:encoding-schemes} summarizes the five different encoding schemes that we consider, which we refer to as APE-G, APE-L, MLP-PE, RoPE-G, and RoPE-L. 

\subsection{Hyperparameters and Training}\label{subsec:dan_training}

\begin{table}
    \centering
    \begin{threeparttable}
    \caption{Names of positional encoding strategies}
    \label{tab:encoding-schemes}
    \begin{tabularx}{\tableWidth\linewidth}{YYY}
        \toprule
        \textbf{Name} & \textbf{Positional Encoding}       & \textbf{Frequency Set}                     \\
        \midrule
        APE-G         & Absolute \eqref{eq:embedding_ape} & Geometric \eqref{eq:frequencies_geometric} \\
        APE-L         & Absolute \eqref{eq:embedding_ape} & Linear \eqref{eq:frequencies_linear}       \\
        MLP           & MLP  \eqref{eq:embedding_mlp}     & N/A                                        \\
        RoPE-G        & Rotary \eqref{eq:rope1}           & Geometric \eqref{eq:frequencies_geometric} \\
        RoPE-L        & Rotary \eqref{eq:rope1}           & Linear \eqref{eq:frequencies_linear}       \\
        \bottomrule
    \end{tabularx}
    \begin{tablenotes}
        \item The five positional encoding schemes were tested in the second stage of the hyperparameter search. Each of them was trained four times with different attention window sizes. A reference to the relevant equation is given for each positional encoding type and frequency set.
    \end{tablenotes}
    \end{threeparttable}
\end{table}

We trained MAST using imitation learning with a mean squared error as the loss function. 
The expert policy was the centralized max-velocity LSAP policy that minimizes the sum of distance-squared. 
Instead of using a pre-generated dataset, we ran simulations during training. 
Examples from these simulations were stored in a replay buffer with a capacity of 20,000 examples.
We roll out 32 parallel simulations for $T=200$ steps during each training epoch. 
TorchRL \cite{Bou23-TorchRLDatadriven} was used for simulation parallelism. 
In half of the simulations, agents used the expert policy; in the other half, agents used the output of the trained model. 
After each simulation step, we stored the 32 observations $\bfo_i(t)$ and expert velocities $\tilde{\bfu}_k(t)$ in the replay buffer and then sampled a mini-batch of 128 such tuples. 
We train using AdamW \cite{Loshchilov2017} with a learning rate of $\eta$ to minimize the mean-squared error of the model output $\bfu_k(t)$ and the expert velocity $\tilde{\bfu}_k(t)$ in each mini-batch. 
After each training epoch, we validate the model by rolling out 32 trajectories with 200 steps, each using the model policy.

\begin{table}[]
    \centering
    \caption{Hyperparameter Search Space}
    \label{tab:dan_hyperparameters1}
    \begin{tabular}{cccc}
         \toprule
         Description & Hyperparameter & Search Space & Optimal Value \\
         \midrule
         \# of layers & $L$ & $\{2,\dots,10\}$ & 4 \\
         \# of heads & $H$ & $\{1,\dots,8\}$ & 4 \\
         head dimension & $d_a$ & $\{16,\dots,128\}$ & 64 \\
         learning rate & $\eta$ & $[\num{e-8},\num{e-2}]$ & \num{e-4} \\
         weight decay & $\lambda$ & $[\num{e-16}, 1]$ & 0 \\
         dropout & $\alpha$ & $[0,1]$ & 0 \\
         \bottomrule
    \end{tabular}
\end{table}

We performed a systematic hyperparameter search consisting of two stages: \emph{(i)} a search over standard hyperparameters and \emph{(ii)} a search over MAST specific hyperparameters.


The search optimized MAST with MLP positional encodings and no attention windowing.
We hypothesized that MAST with RoPE encodings and attention windowing would perform favorably.
We intentionally optimize the hyperparameters for MLP-PE to minimize the impact of our bias.

We ran 400 trials in a Hyperband search~\cite{Li17-HyperbandNovel} with a reduction factor of 4, a minimum of 10 epochs, and a maximum of 100 epochs.
The parameters for each trial were sampled from the above ranges using the tree-structured Parzen estimator (TPE)~\cite{Bergstra11-Algorithms}. 
We used the Optuna~\cite{Akiba19-OptunaNextgeneration} implementation of Hyperband search and TPE. 
We chose the model that maximized the average success rate from \eqref{eq:success_rate}. 
We rounded the hyperparameters to arrive at $L=4$ layers, $H=4$ heads, $d_a=64$ head dimension, learning rate $\eta=1e-4$, weight decay $\lambda=0$, and dropout of $\alpha=0.0$.
We used these hyperparameters for all models in future experiments.

\damian{Having found suitable basic hyperparameters, we trained 75 different MAST models. We vary the MAST variant from Table~\ref{tab:transformer-variants}, the positional encoding scheme, frequency sets, and attention window sizes. }
Each run employed a positional encoding scheme, which could be either Absolute Positional Encoding (APE), Multi-Layer Perceptron Positional Encoding (MLP-PE), or Rotary Positional Encoding (RoPE).
The frequencies $\omega_k$ for RoPE and APE were sampled geometrically, as described in Equation \eqref{eq:frequencies_geometric}, or linearly, as outlined in Equation \eqref{eq:frequencies_linear}, with a base period of $\baseWavelength=\SIkm{1}$.
The attention window radius was set to \( R_{\mathrm{att}} \in \{\infty, 250, 500, 1000\} \), where \( R_{\mathrm{att}} = \infty \) indicates that no windowing was applied.
We train each model with $N=100$ agents for 500 epochs.

\subsection{Evaluation}\label{subsec:dan_evaluation}

\damian{
Reorder:
\begin{enumerate}
    \item Compare against baselines.
    \item Generalization to different scenarios.
    \item Robustness to delays.
    \item Compare with and with connectivity masking --- how this makes delays worse.
    \item Choice of attention window size, positional encoding.
\end{enumerate}

\red{We evaluate the models in several ways.}
First, in Section~\ref{subsec:dan_gridsearch}, we focus on the impact of component masking, attention windowing, and positional encodings on size generalization under centralized execution.
Second, in Section~\ref{subsec:dan_decentralized_execution}, we discuss the impact of communication delays on the decentralized execution performance of MAST-M and MAST-L.
Finally, in Section~\ref{subsec:dan_ood}, we study how well the MAST architecture performs on out-of-distribution scenarios.

}
\subsubsection{Comparison to baselines}\label{subsec:dan_baselines}

\begin{figure}
    \centering
    \includegraphics[width=\figwidth\linewidth]{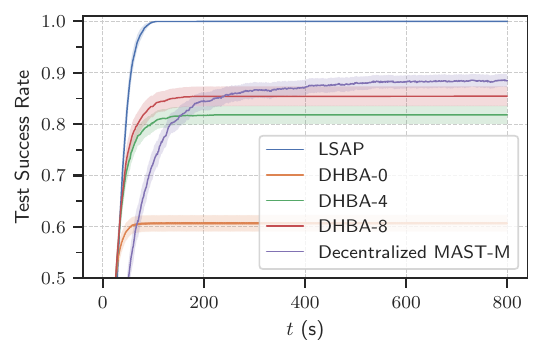}
    \caption{Plots of mean success rate over time across 100 test simulations. 
    Decentrally executed MAST-M at a communication delay of $\tau / \Delta t = 1.0$ is compared to baseline policies. 
    The baselines are explained in Sections~\ref{subsec:dan_centralized_policies}~and~\ref{subsec:dan_decentralized_policies}. 
    MAST-M takes longer to converge, eventually outperforming all the decentralized DHBA policies.
    }
    \label{fig:compare-baseline}
\end{figure}

In Sections \ref{subsec:dan_centralized_policies} and \ref{subsec:dan_decentralized_policies}, we discussed baseline centralized and decentralized policies that solve the assignment and navigation problem by formulating a linear sum assignment problem. We referred to these policies as LSAP and DHBA-k, where $k$ is the number of hops in $\calG$ from which information is aggregated to run the decentralized Hungarian algorithm. 
Figure~\ref{fig:compare-baseline} compares MAST-M against these baseline policies. 
For this figure, MAST-M was executed in a decentralized manner with a communication delay of $\tau / \Delta t = 1$. 
For comparison, we plot the success rate of DHBA for $k = 0, 4, 8$. 
DHBA performance improves as the number of hops increases, but the returns are diminishing, and performance does not improve further for $k > 8$. 
We do not time-delay inputs to DHBA in our experiments, unlike when executing MAST-M. 
Despite this disadvantage, MAST-M achieves a higher success rate than DHBA. 
MAST-M takes longer to converge, having a lower success rate for the first 200 seconds.

\subsubsection{Out of distribution scenarios}\label{subsec:dan_ood}

\begin{figure}
    \centering
    \includegraphics[width=\figwidth\linewidth]{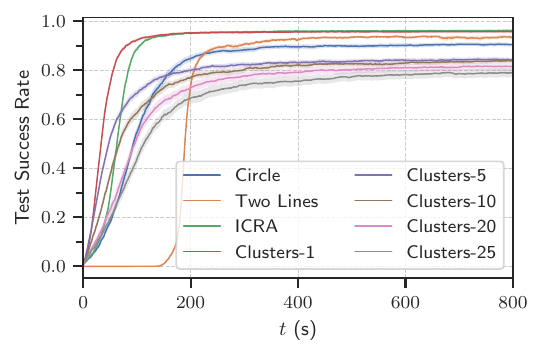}
    \caption{Test success rate of MAST-M in different scenarios. The initial conditions of `clusters-1', `clusters-5', and `clusters-10' are similar to those encountered in training. The remaining scenarios are not sampled in the training data.}
    \label{fig:scenarios}
\end{figure}

\begin{figure*}
    \centering
    \includegraphics[width=\figwidthDouble\linewidth]{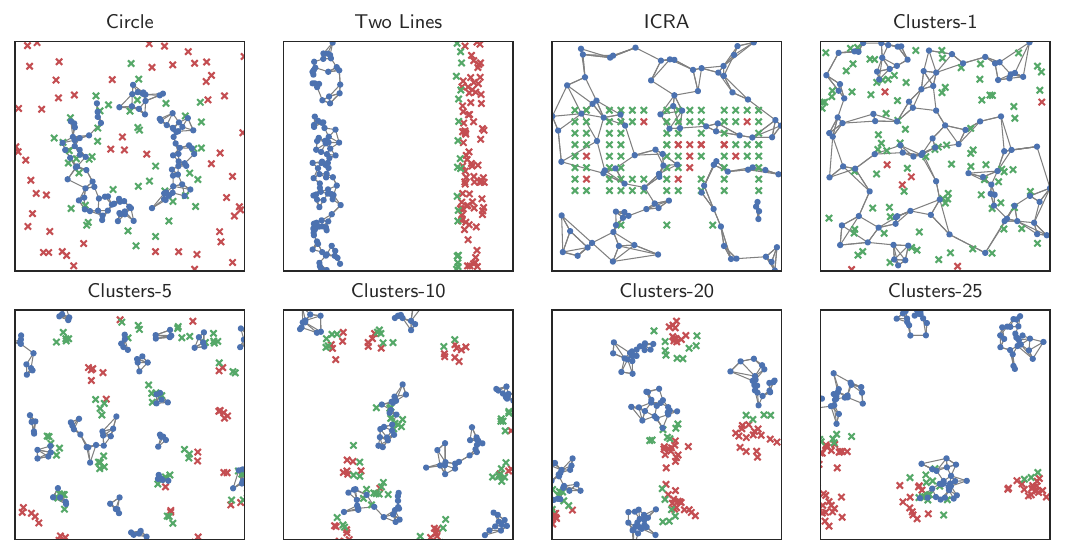}
    \caption{
        Examples of different scenarios, as described in Section~\ref{subsec:dan_ood}. 
        Blue circles represent agents, green \green{$\bm{\times}$}'s represent goals observed by at least one agent, and red \red{$\bm{\times}$}'s represent goals unobserved by any agent.
        Lines represent the communication graph $\calG$.
        `Clusters-1', `Clusters-5', and `Clusters-10' are the only ones that are similarly distributed to the training data. All other scenarios are out of distribution.
    }
    \label{fig:scenarios-initialization}
\end{figure*}

Generalizability to unseen scenarios is a necessary property of a practical solution to DAN.
Although Section~\ref{subsec:simulation_environment} explains this in more depth, recall that, during training, the agents and goals are initialized in clusters of 1, 5, or 10. 
To test out-of-distribution generalizability, we introduce several new randomly generated scenarios.
Figure~\ref{fig:scenarios-initialization} shows example initial conditions for each scenario, and we describe each below.
In the \emph{`circle'} scenario, goals are uniformly distributed, and agents are uniformly randomly arranged in a hollow disk of inner radius $3W/16$ and outer radius $5W/16$. 
In the {\emph{`two lines'}} scenario, agent positions are randomly arranged in a line so that the agent positions are uniformly distributed with $\bfp_i(0) \sim U(-W/4,-3W/8) \times U(-W/2, W/2)$. Similarly, the goal positions are sampled from $\bfg_i \sim U(3W/8,W/4) \times U(-W/2,W/2)$. 
In the \emph{`ICRA'} scenario, the initial agent positions are uniformly distributed within the environment, while the goals are arranged to spell out `ICRA.' 
Finally, {\emph{`cluster-k`}} for $k \in \{1,5,10,20,25\}$ are a family of scenarios where the agents and goals are each randomly arranged in clusters of size $k$ -- as described in Section~\ref{subsec:simulation_environment}, with $\rho_\mathrm{agent} = \rho_\mathrm{goal} = k$.
For $k \in \{1, 5, 10\}$, the cluster-k scenarios are very similar to the training distribution.

MAST delivers a high success rate for all scenarios, but there is some variability between scenarios. 
For example, there is a clear trend in performance for the clustered family of scenarios: the higher the cluster size, the lower the success rate.
The mean success rate for the `clusters-1` scenario is over 0.9, but for `clusters-25` it is below 0.8. 
This deterioration in performance can be explained by the communication graph becoming increasingly disconnected as the cluster size increases. 
Figure~\ref{fig:scenarios-initialization} exemplifies this: in the `clusters-1' scenario, the communication graph is more connected than in scenarios with larger clusters. 
This also explains why the `circle,' `two lines,' and `ICRA' scenarios perform significantly better than the clustered scenarios for $k > 1$.
Overall, the final success rate and convergence rate highly depend on the connectivity of $\calG$.

\subsubsection{Decentralized execution}\label{subsec:dan_decentralized_execution}

\begin{figure*}
    \centering
    \includegraphics[width=\figwidthDouble\linewidth]{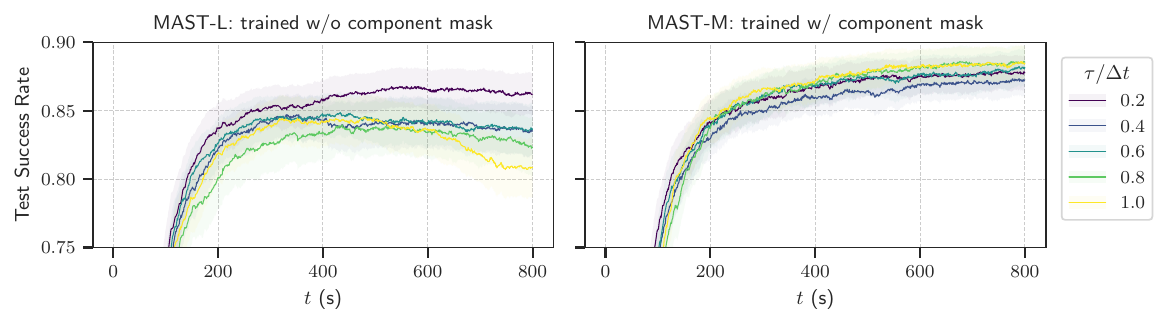}
    \caption{
    Decentralized execution performance of MAST, centrally trained without component masking (MAST-L, left) and trained with component masking (MAST-M, right). Both models use RoPE-G positional encodings and attention window radius $R_\mathrm{att} = 250$.
    Each line represents the mean success rate over 100 simulations for each combination of time-delay, $\tau$, and model. The width of the shaded region around each line is a 95\% confidence interval. Color indicates the time delay, $\tau / \Delta t$, expressed relative to the MAST update period. 
    To improve readability, we only plot trials with a time delay that is a multiple of $0.2$ seconds. MAST-L performance deteriorates significantly as the time delay increases. 
    Component masking greatly benefits MAST-M, which can maintain performance despite time delays. 
    In fact, for MAST-M, the mean success rate is higher for $\tau\in \{ 0.6, 0.8, 1.0 \}$ than for $\tau / \Delta t = 0.2$, although they are within each other's confidence intervals.
    }
    \label{fig:delay-over-time}
\end{figure*}

\begin{figure}
    \centering
    \includegraphics[width=\figwidth\linewidth]{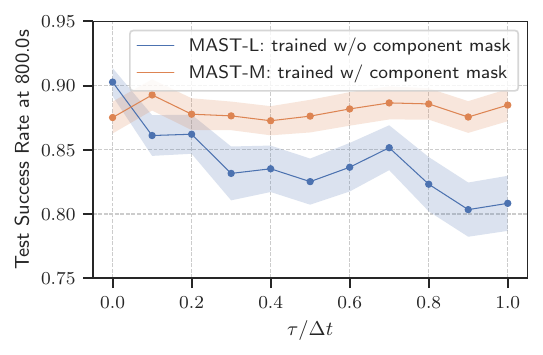}
    \caption{
    Decentralized execution performance of MAST centrally trained without component masking \blue{(MAST-L, blue)} and trained with component masking \orange{(MAST-M, orange)}. The plot shows the relationship between the communication delay, $\tau / \Delta t$, and the terminal success rate at $t = 800$ seconds. 
    The communication delay is expressed relative to the MAST update period.    
    The mean value across 100 simulations is reported, along with a 95\% confidence interval. Reference centralized execution results are at $\tau / \Delta t = 0$. MAST-L performs better than MAST-M with centralized execution, but performance deteriorates when delays are added. The MAST-M success rate is unaffected by decentralized execution and increasing delay.
    }
    \label{fig:delay-terminal}
\end{figure}

After centralized training, MAST-M and MAST-L can be executed on a decentralized multi-agent system, as described in Section~\ref{subsec:decentralized_inference}.
We evaluate the decentralized execution performance of these models by simulating communication using Algorithm~\ref{alg:communication}.
We evaluate the performance of both models for varying communication delays, $\tau / \Delta t \in \{0.1, 0.2, \dots, 1.0\}$, which we express in terms of the MAST controller update period.
We perform 100 simulations for 800 seconds for each model and time-delay combination, with $N = 100$ agents.
Note that MAST-C is not tested in this way because it is incapable of decentralized execution due to its reliance on global knowledge of the goal positions.

The results of these experiments are shown in Figures~\ref{fig:delay-over-time} and~\ref{fig:delay-terminal}. 
The former shows the success rate over time, while the latter focuses on the terminal success rate. 
MAST-L is significantly affected by communication delays, with performance negatively correlated with delay duration.
In Figure~\ref{fig:delay-over-time}, MAST-L has much more variability in the mean success rate and has larger error bands. 
Figure~\ref{fig:delay-terminal} shows that the MAST-L success rate degrades from 0.9 with centralized execution down to just over 0.8 with delays of $\tau  = 1$ second.
On the other hand, MAST-M is much more resilient to delays. 
In Figure~\ref{fig:delay-over-time}, the differences between tests at different delays do not appear significant. 
The highest mean performance is achieved by $\tau / \Delta t = 1.0$ and $\tau / \Delta t = 0.8$, with lower delays, such as $\tau / \Delta t = 0.2$, leading to inferior performance. 
This is further evidenced by Figure~\ref{fig:delay-terminal}, which shows that time delays up to 1 do not significantly affect the MAST-M terminal success rate. Moreover, there does not appear to be a significant difference in success rate between centralized execution ($\tau / \Delta t = 0$) and decentralized execution ($\tau / \Delta t > 0$).

\begin{figure}
    \centering
    \includegraphics[width=\figwidth\linewidth]{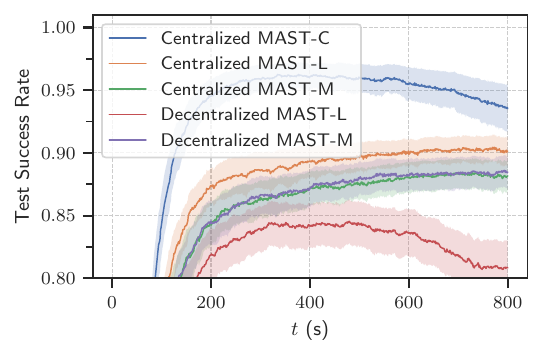}
    \caption{
    Plots of mean success rate over time across 100 test simulations for different variants of MAST. MAST-C is executed centrally. The plots show include centralized and decentralized execution for MAST-L and MAST-M.
    During decentralized execution, we use a communication delay of $\tau / \Delta t = 1$.    
    Of the centralized execution experiments, MAST-C has the highest success rate, followed by MAST-L and MAST-M. 
    The success rate for MAST-M does not change when executed in a decentralized manner. 
    This is unlike MAST-L, which has the lowest success rate under decentralized execution.
    }
    \label{fig:compare-transformer}
\end{figure}

\fgref{fig:compare-transformer} compares different variants of MAST executed centrally or decentralized.
For decentralized execution of MAST, the figure reports results with a communication delay of $\tau / \Delta t = 1$.
When execution is centralized, MAST-C performs the best in terms of success rate, followed by MAST-L and then MAST-M. 
When execution is decentralized, the performance of MAST-L deteriorates to below MAST-M.
The MAST-M success rate does not significantly vary between centralized and decentralized execution.

For centralized execution, the differences in success rates between MAST variants are unsurprising.
MAST-C can leverage global goal position information to improve the success rate over MAST-L. Similarly, unlike MAST-M, MAST-L is not constrained by component masking, allowing attention between agents that are in distinct connected components on $\calG$. 
MAST-L can leverage this information to improve the success rate. 
However, this becomes a disadvantage under decentralized execution when this information is no longer available and, therefore, the input data distribution is different from that in training.
This distributional shift between the centralized training and decentralized execution is absent for MAST-M. 
Adding delays during decentralized execution also introduces a distributional shift but does not significantly affect performance, as evidenced by Figure~\ref{fig:delay-terminal}. 
Probably, information from far-away agents is unimportant for determining local policies, which limits the impact of delayed information about many hop neighbors.

\vspace{10em}


\fgref{fig:masked-comparison} shows the performance of the MAST-M model for each encoding scheme and attention window radius combination. First, we consider the relative performance of all the models when the number of agents is fixed at $N = 100$.
There are two apparent trends:
\emph{(i)}~RoPE consistently outperforms the other positional encoding strategies and 
\emph{(ii)}~positional encodings with geometric frequency sets are less affected by changes in window size than linear frequency sets.
RoPE-G has consistently high performance regardless of attention window size. For every choice of window size, APE-G performs worse than RoPE-G.
Nevertheless, like RoPE-G, it is robust to changes in attention window size. 
Attention window size has a larger impact on the performance of positional encodings with linear frequencies: RoPE-L and APE-L. 
The two perform similarly for $R_\mathrm{att} = 1000$, but in all other cases, RoPE-L dominates APE-L.
The performance of MLP-PE varies for different attention window sizes, with better performance for $R_\mathrm{att} \in \{500, \infty\}$ than $R_\mathrm{att} \in \{250, 1000\}$.

Generalizability to larger teams of agents is depicted by the bottom row of \fgref{fig:masked-comparison}. RoPE-G consistently achieves the highest success rate, except for $R_\mathrm{att} = 250$, where RoPE-L performs marginally better. 
RoPE's generalizability is affected by the attention window; overall, models with larger window sizes have a more pronounced deterioration in performance. 
RoPE-G and RoPE-L offer the best trade-off between success rate and generalizability when $R_\mathrm{att} = 250$. MLP-PE and APE-G generalizability is largely unaffected by the attention window size. 
Their performance deteriorates as the number of agents increases, although APE-G's performance deteriorates much more. 
APE-L outperforms APE-G for $N > 100$. Unlike other positional encodings, APE-L's performance does not decrease monotonically with $N$.

Figure~\ref{fig:local-comparison} shows the performance of the MAST-L model for each encoding scheme and attention window radius combination. First, we focus on the case for $N = 100$ agents. 
RoPE-G and RoPE-L perform best for $R_\mathrm{att} \le 1000$, but as the window size increases, the performance of RoPE-G improves while the performance of RoPE-L deteriorates.
Similarly, APE-G tends to outperform APE-L and achieves a relatively consistent success rate, regardless of the attention window. 
MLP-PE achieves a consistent success rate, regardless of attention window size. In fact, for $R_\mathrm{att} = \infty$, MLP-PE achieves the second highest success rate, after RoPE-G. Looking at just $N=100$, RoPE-G is the preferable positional encoding scheme because it performs best for larger attention radii and is only marginally worse than RoPE-L for smaller radii.

RoPE-G, RoPE-L, and MLP-PE obtain the best generalization performance for large agent teams when the attention window is $R_\mathrm{att} = 250$. The generalization performance of all three deteriorates for $R_\mathrm{att} \in \{250, 500\}$. However, while the performance of RoPE continues to deteriorate when attention windowing is removed, MLP-PE does not follow the same trend. Instead, MLP-PE generalizes similarly when $R_\mathrm{att} = \infty$ and when $R_\mathrm{att} = 250$. 
MLP-PE appears less sensitive to the choice of attention window size than APE and RoPE.
Except for MLP-PE, the generalization performance of MAST-L is more sensitive to the choice of attention window radius than MAST-M. 
For MAST-L with $R_\mathrm{att} \ge 1000$ and $N = 1000$, only one positional encoding performs above a 0.6 success rate. 
Conversely, for MAST-M with $R_\mathrm{att} \ge 1000$ and $N = 1000$, all but one positional encoding achieved a success rate above 0.6. 

We hypothesize the following explanation for this difference in sensitivity to the attention window size.
Recall that the only difference between MAST-L and MAST-M is that the latter employs component masking.
Including component masking in MAST makes the architecture more local, as the likelihood that two agents are in the same connected component decreases with distance.
Hence, there is a lower likelihood of performance-detrimental phenomena such as aliasing, described in Section~\ref{subsec:positional_encoding_frequencies}.

\subsection{Discussion}

In the first part of Section~\ref{subsec:dan_evaluation}, we emphasize the significance of selecting both the type of positional encoding and the size of the attention window for performance in decentralized assignment and navigation tasks. 
Learnable positional encodings (MLP-PE), demonstrate reasonable performance during centralized execution with MAST-C. 
They also maintain consistent performance in semi-decentralized and decentralized settings with MAST-L and MAST-M. 
However, RoPE is more effective at leveraging positional information. 
On its own, RoPE cannot generalize to scenarios with more agents than during training. 
By utilizing attention windowing or component masking, we can make the architecture more local, which enables generalizability to large training teams at a slight performance cost.

The primary advantage of component masking is its impact on decentralized execution performance. 
Section~\ref{subsec:dan_decentralized_execution} provides our experimental results with decentralized execution.
It is clear from Figure~\ref{fig:delay-terminal} that component masking negatively impacts performance during centralized execution. 
However, during decentralized execution, the model trained using component masking (MAST-M) is unaffected by communication delays of up to one second.
This is further evidenced by MAST-M's high and consistent performance over MAST-L in Figure~\ref{fig:delay-over-time}. 
In our opinion, the most probable explanation is that component masking lowers the distributional gap between centralized training and decentralized execution.

The capability of decentralized execution, robustness to communication delays, generalizability to out-of-distribution scenarios, and generalizability to larger teams of agents suggest that MAST-M is a viable approach to decentralized assignment and navigation. 
However, we would like to highlight some limitations in our analysis. 
Each model was only trained once, so the estimated confidence intervals do not account for training variability.
Additionally, our experiments always use the k-nearest neighbor communication model. 
The approach generalizes well to execution with delayed information, but transferability to different communication models is not fully explored.
This is partially addressed in the coverage control experiments, where we use a distance-based communication graph.

\section{Coverage Control}\label{sec:coverage_control}
\subsection{Problem Formulation}

Coverage control is the problem of commanding a team of robots to provide sensor coverage to areas of variable importance.
The specifications of the task determine these areas of importance.
We consider the coverage problem on the plane $\calX \subset \reals^2$ on which we deploy a team of $N$ homogenous, holonomic agents indexed by $\calV = \{1, \dots, N\}$. 
As defined in Section~\ref{sec:decentralized-robotics}, consider positions $\bfP(t)$, control inputs $\bfU(t)$, and first-order integrator dynamics from Equations \eqref{eq:agent_position}, \eqref{eq:control}, and \eqref{eq:dynamics}.
Once again, we assume the existence of a dynamic communication graph $\calG(t) = (\calV, \calE(t))$, but consider a different structure.
This section considers a communication range $R_{c}$.
Any pair of robots $i,j \in \calV$ is within communication range when $\lVert \bfp_i(t) - \bfp_j(t) \rVert < R_{c}$ and therefore we have an edge $(i,j) \in \calE(t)$.

The aforementioned areas of interest define the \textit{importance density function} (IDF).
The IDF, expressed as a function $\Phi: \calX \rightarrow \reals^{+}$, assigns an importance value to each point in the environment.
Following \cite{Agarwal2024, CortesMKB02}, the coverage cost $\calJ$ is defined in Equation~\eqref{eq:coverage_cost} and is determined by the positions of robots in the environment and the IDF:
\begin{equation} \label{eq:coverage_cost}
    \calJ(\bfP(t)) = \int_{\bfq \in \calX} \min_{i\in\calV} f(\lVert \bfp_i(t) - q \rVert) \Phi(\bfq) d\bfq
\end{equation}
where $f$ is a non-decreasing function such as $f(x) = x^2$.

Computing Equation~\eqref{eq:coverage_cost} directly for large numbers of agents is computationally expensive due to the minimization problem inside the integral.
Instead, assuming that robots have distinct positions and $f(x) = x^2$, consider a partition of the position-space $\calX = \cup_i P_i $ into Voronoi cells \cite{CortesMKB02}:
\begin{equation} \label{eq:voronoi_partition}
    P_i = \{ \bfq \in \calX \: \mid \: \lVert \bfp_i - q \rVert \leq \lVert \bfp_j - \bfq \rVert, \forall \; j \in \calV\}
\end{equation}
for all $i \in \calV$. The Voronoi cell $P_i$ is a subset of $\calX$ where all points are closest to the position of the $i$th agent.
The sets $P_i$ disjointly partition $\calX$, so we can rewrite the coverage cost as
\begin{equation} \label{eq:coverage_cost_voronoi}
    \calJ(\bfP(t)) = \sum_{i=1}^N \int_{\bfq \in P_i} \lVert \bfp_i(t) - q \rVert^2 \Phi(\bfq) d\bfq.
\end{equation}
The partitions can be efficiently computed using Lloyd's algorithm \cite{Lloyd82-LeastSquares}.

There are several approaches to coverage control in the literature that build on Lloyd's algorithm \cite{Cortes05-Spatiallydistributed, CortesMKB02, Rudolph21-RangeLimited, Bai22-AdaptiveMultiAgent, Carron20-ModelPredictive}.
Typically, these are broadly categorized by their framing of the problem and their accompanying assumptions.
Cort\'es et al. \cite{CortesMKB02,Cortes05-Spatiallydistributed} were the first to demonstrate a gradient descent algorithm based on Lloyd's algorithm that could solve the coverage problem using both centralized and decentralized Centroidal Voronoi Tesselation (CVT); abbreviated as C-CVT and D-CVT, respectively.
However, the authors assume that the density function is known \textit{a priori}.
Rudolph et al. \cite{Rudolph21-RangeLimited} consider cases of heterogeneous robot teams with different sensing capabilities and where the density function is initially unexplored.
The authors propose a fusion between a local implementation of Lloyd's algorithm and a higher-level distribution controller that leverages the different sensing capabilities of individual robots.
\cite{Bai22-AdaptiveMultiAgent} and \cite{Carron20-ModelPredictive} respectively tackle coverage control under obstacle constraints and dynamics.
The latter takes a model predictive approach to solve coverage control with constraints on states and inputs.

Imitation learning-based methods have relied on Lloyd's algorithm as well \cite{Agarwal2024, Gosrich2022}.
As before, coverage control can be framed in numerous ways.
Gosrich et al. \cite{Gosrich2022} proposed a perception action communication (PAC) architecture that uses a CNN for perception and a GNN for communication.
They consider limited-range IDF sensing but assume full connectivity between agents.
Their method outperforms the centralized implementation of Lloyd's algorithm and demonstrates transferability to non-convex environments.
More recently, Agarwal et al. \cite{Agarwal2024} introduced LPAC-K3, a similar architecture to \cite{Gosrich2022} that also assumes that the IDF is unknown at $t = 0$.
However, they consider the case where communication only occurs between an agent and its proximal neighbors.
To compare our proposed architecture, we consider three variants of CVT and LPAC-K3 \cite{Agarwal2024}.
Table \ref{tab:coverage-baselines} summarizes the information access of each method.

\subsubsection{Clairvoyant CVT}
Clairvoyant CVT is an implementation of CVT that has complete knowledge of the IDF and the robot positions.
Notably, this algorithm is not guaranteed to converge to a global minimum.
Nevertheless, it performs well empirically and serves as the expert with which we train our proposed architecture with imitation learning.

\subsubsection{Centralized CVT}
In C-CVT, agent positions are known globally, but access to the IDF is restricted to areas the team has uncovered through exploration.
In other words, the policy has access to the portion of the IDF that any agent has seen at any time up to the current time step.
Each agent has a restricted sensor FOV with which to uncover the IDF.

\subsubsection{Decentralized CVT}
In D-CVT, agents are restricted in their ability to sense the IDF and have limited information about the positions of their neighbors, as determined by the communication radius.
Agents can only compute the Voronoi cells using the positions of their neighbors, thereby rendering the computed partitioning incomplete unless the communication graph is complete.

\subsubsection{LPAC-K3}
Like D-CVT, agents are restricted in their ability to sense the IDF and have limited information about the positions of their neighbors. Unlike D-CVT, LPAC-K3 uses only local observations from the current timestep. The model uses a GNN to disseminate learned embeddings across the network. LPAC-K3 is the current state-of-the-art learning-based model that has demonstrated excellent performance compared to C-CVT and D-CVT.

\subsection{Perception Model}

In our simulations, each agent only has access to local information gathered from its surroundings, similarly to \cite{Agarwal2024}.
Separate from the communication radius $R_c = \SI{256}{\meter}$, the agents have a limited sensor field of view $(\SI{64}{\meter} \times \SI{64}{\meter})$ centered at the agent's current position.
Although agents uncover new information about the environment through sensing, they can use previously seen information as long as it is within the local map $(\SI{256}{\meter} \times \SI{256}{\meter})$, also centered at the agent's current position.
The communication radius and the sensor field of view are properties of the agent hardware and are, therefore, fixed for all experiments.
From the environment, each agent makes the observation $\bfo_i(t) \in \reals^{4 \times d_o \times d_o}$, which is a four-channel image.
The channels of $\bfo_i(t)$ are $\bfC_{\Phi}^i, \bfC_{B}^i, \bfC_x^i, \bfC_y^i$.
Each channel is a \emph{map} of the environment near the $i$th agent.
The pixels of $\bfC_{\Phi}^i$ represent the values of sensed IDF.
$\bfC_{B}^i$ represents the bounds of the simulation environment, taking unit values for pixels corresponding to positions outside the bounds and zero otherwise.
$\bfC_x^i$ and $\bfC_y^i$ are maps indicating the relative spatial positions of neighboring agents that fall within the communication range.
Each channel is initially sampled at a resolution of $(256 \times 256)$ but is scaled down to $(32 \times 32)$.

We use a CNN as the perception model, since the observations $\bfo_i(t)$ are images. The CNN processes the observations and produces a vector embedding  $\bfx_i(t) \in \reals^{32}$.
Following \cite{Agarwal2024}, the perception model has three layers, 32 channels, a kernel size of three, and a final output dimension of 32.
The CNN uses batch normalization and leaky-ReLU pointwise nonlinearities.
The output of the final convolutional layer is flattened and projected linearly to $\reals^{32}$.
Although we use the same architecture and hyperparameters as \cite{Agarwal2024}, we did not use a pre-trained perception model, training the LPAC stack end-to-end.

\begin{figure}
    \centering
    \includegraphics[width=\linewidth, trim=300px 100px 300px 100px, clip]{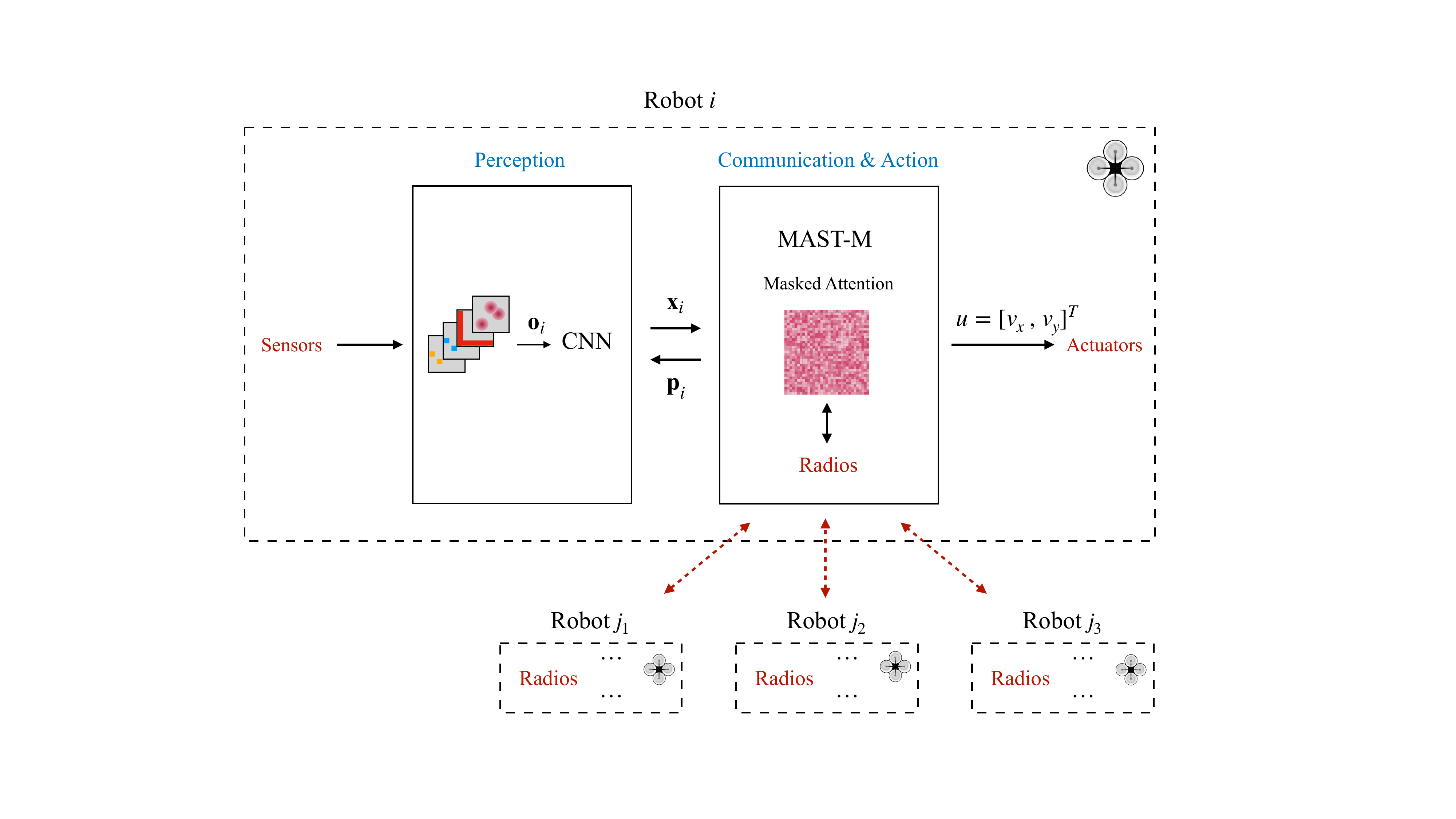}
    \caption{
        \fred{redundant}
        The Perception-Action-Communication loop is parameterized by a convolutional neural network (CNN) and MAST-M.
        The CNN processes local observations to produce a vector embedding.
        This embedding is shared with neighboring robots, each processing the set of input embeddings from their respective neighbors using MAST-M.
        The model outputs the control action to be taken by the respective agent.
    }
    \label{fig:coverage-lpac-transformer}
\end{figure}

\subsection{Training}

We use imitation learning to train MAST-M for the coverage control task.
Using the Clairvoyant CVT policy, we generate a dataset with 100,000 examples.
Each example contains the communication graph, raw observations, positions, and actions for all $N$ agents at the recorded timestep.
For each environment in the dataset, the IDF was prepared by uniformly placing $F = 32$ bivariate Gaussians in the planar space of dimension $1024 \times \SI{1024}{\meter\squared}$.
The initial positions of each of the $N = 32$ agents are also uniformly distributed across the environment.
Each expert rollout was sampled every 5 time steps to reduce the temporal dependence between samples.

We conducted a Bayesian hyperparameter search over MAST and optimizer hyperparameters.
The final model has $L = 8$ layers, $H = 8$ heads, $d_a = 32$ head dimension, learning rate $\eta = 3.3e-4$, weight decay $\lambda = 2e-9$ and a dropout of $\alpha = 0.0$.
Next, a grid search was conducted over the attention radius $R_{att} \in \{256, 512, 1024\}$ with and without component masking.
The final model uses $R_{att} = \SI{256}{\meter}$ as well as RoPE-G and was trained in the MAST-M configuration, where both the local mask, $M_{ij}^P$, and the component mask, $M_{ij}^C$, are used.
Each model was trained for 100 epochs with a batch size 256 and the AdamW optimizer.

\subsection{Evaluation}

We evaluate MAST-M in different scenarios to demonstrate the effectiveness of the model.
Section \ref{subsec:coverage_baselines} compares rollouts of all policies on 100 randomly generated in-distribution environments.
Section~\ref{subsec:coverage_env_size} and Section~\ref{subsec:coverage_scale} explore the model's performance when the environment size or agent and feature densities are varied.
Finally, we include Section \ref{subsec:coverage_real_world} to show how MAST-M performs when the IDF is created from semantic information in the environment rather than randomly generated.

For scenarios where the environments are randomly generated, we construct the IDF by drawing center positions of $F$ bivariate Gaussians from $U(0, \ell_e) \times U(0, \ell_e)$ where $\ell_e$ is the length of one side of the square environment.
Each Gaussian is truncated to 2 standard deviations to limit the detection of features by robots far from their centers.
The peak amplitude of each Gaussian is randomly sampled from $U(0.6, 1.0)$.
The initial positions of agents are sampled according to $\bfp_i(0) \sim U(0, \ell_e) \times U(0, \ell_e)$.
Each control policy is evaluated in environments with the same initial conditions to ensure a fair comparison.

\subsubsection{Comparison with the Baselines}\label{subsec:coverage_baselines}

\begin{figure}
    \centering
    \includegraphics[width=\linewidth]{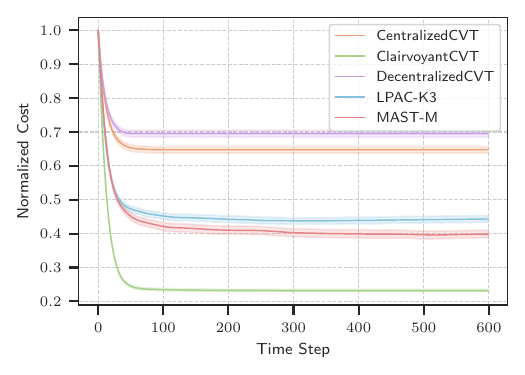}
    \caption{
        The normalized coverage cost of baseline controllers compared to MAST-M over 100 environments.
        The episode length is 600 timesteps.
        The cost of each controller is normalized by its initial cost at the first timestep.
        The solid lines show each controller's mean coverage cost across all rollouts, and the shaded band represents the \SI{95}{\percent} confidence interval of the mean.
        MAST-M was trained with a component mask that was informed by the emergent communication graph ($R_{c} = 256$) and with an attention window radius, $R_{att} = 256$.
        The transformer-based model outperforms both the CVT baselines and the GNN-based LPAC-K3.
    }
    \label{fig:coverage-basic-eval}
\end{figure}

We compare MAST-M against four different baselines, summarized in Table~\ref{tab:coverage-baselines}.
Each policy is rolled out over 100 randomly generated environments for 600 time steps.
Empirically, all controllers converge within this time frame.
The environment has a width of \SI{1024}{\meter} and is initialized with 32 agents and 32 features.
Figure \ref{fig:coverage-basic-eval} shows the performance of each policy over time.
For each policy, the dark line is the mean coverage cost across environments for each time step, while the bands show the standard deviation.
The coverage cost of each controller is normalized by the initial coverage cost at $t = 0$ for each rollout.
Regardless of the policy used, the agents have a fixed communication radius $R_{c} = 256$ and a sensor field of view of $(64 \times \SI{64}{\meter\squared})$

\begin{table}
    \centering
    \begin{threeparttable}
        \caption{Coverage Control Policies}
        \label{tab:coverage-baselines}
        \begin{tabularx}{\tableWidth\linewidth}{YYY}
            \toprule
            \textbf{Policy}            & \textbf{IDF Access} & \textbf{Position Access} \\
            \midrule
            Clairvoyant CVT            & Full                & Full                     \\
            Centralized CVT            & Partial             & Full                     \\
            Decentralized CVT          & Partial             & Partial                  \\
            LPAC-K3 \cite{Agarwal2024} & Partial             & Partial                  \\
            \midrule
            MAST-M (Ours)              & Partial             & Partial                  \\
            \bottomrule
        \end{tabularx}
        \begin{tablenotes}
            \item The four baselines for comparison and MAST-M.
            The CVT approaches differ in their information access.
            LPAC-K3 is a state-of-the-art imitation learning approach to coverage control that uses a GNN to leverage the communication graph structure.
        \end{tablenotes}
    \end{threeparttable}
\end{table}

\subsubsection{Increasing the Environment Size}\label{subsec:coverage_env_size}

\begin{figure}
    \centering
    \includegraphics[width=\linewidth]{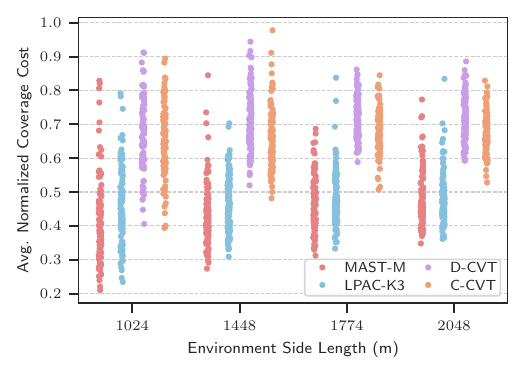}
    \caption{Performance of MAST-M as the scale of the environment is increased.
        The density of agents and features remains constant as the environment size is increased.
        The strip plot shows the distribution of average normalized coverage cost over 100 environments for each policy.
        MAST-M decisively outperforms the D-CVT and C-CVT baselines while maintaining a consistent average coverage cost as the environment size is increased.
        The transformer with an attention window limited to $R_{att} = 256$ can match the performance of LPAC with a 15-hop reach (5 layers, three hops per layer).%
    \label{fig:coverage-strip}}
\end{figure}

It is crucial that a decentralized policy scales efficiently with the dimensions of the environment.
In these experiments, we evaluate the performance of MAST-M against the decentralized CVT and LPAC-K3 policies as the size of the environment changes.
We evaluate each policy over $\ell_{env} \in \{1024, 1448, 1774, 2048\} \; \unit{\meter}$ where the total area is $A_{env} = \ell_{env} \times \ell_{env}$.
The densities of agents and Gaussian features remain constant. 
For $\ell_{env} = 1024$ we have $N = 32$, $F = 32$ and for $\ell_{env} = 2048$ we have $N = 128$, $F = 128$). 
Each policy is evaluated 100 times at each environment size.
The  strip plot in Figure~\ref{fig:coverage-strip} shows the distribution of the average normalized coverage cost for 600 time steps over these 100 environments compared against D-CVT and LPAC-K3, respectively.
There is a significant performance improvement from the D-CVT baseline to MAST-M.
The distribution overlap decreases as $\ell_{env}$ increases.
MAST-M remains consistent and can generalize to larger-sized scenarios after training on small examples.
Although MAST-M significantly outperforms LPAC-K3 for $\ell_{env} = 1024$, this gap closes as the environment size increases.
Their distributions are almost identical for $\ell_{env} = 2048$.

%

\begin{figure}
    \centering
    \includegraphics[width=\linewidth]{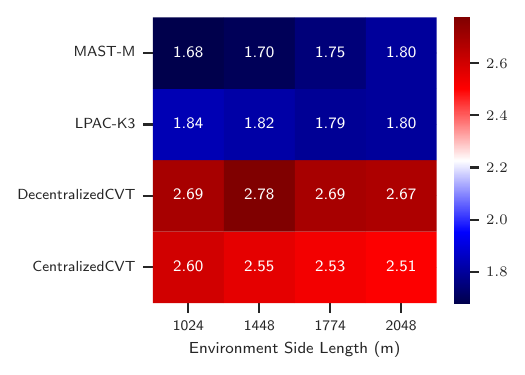}
    \caption{
        \fred{redundant}
        Policy comparison for different environment side lengths.
        The values displayed are the ratio between the average normalized cost of the respective policy and the clairvoyant expert.
        MAST-M outperforms all baselines and matches the performance of LPAC-K3 for an environment side length of \SI{2048}{\meter}.
    }
    \label{fig:coverage-env-heatmap}
\end{figure}

The policies are compared against the clairvoyant expert in Figure~\ref{fig:coverage-env-heatmap}.
Each value represents the ratio of the average normalized coverage costs of the given controller and the expert, respectively.
MAST-M is closer to the performance of the clairvoyant expert than the baselines.
When $\ell_{env} = 2048$, MAST-M and LPAC-K3 achieve the same performance relative to the expert.
Although MAST-M outperforms LPAC-K3 for $\ell_{env} < 2048$, its performance negatively correlates with environment side length, unlike LPAC-K3.

\subsubsection{Agent and Feature Density}\label{subsec:coverage_scale}

\begin{figure*}[htpb]
    \centering
    \subfloat{\includegraphics[width=0.45\textwidth]{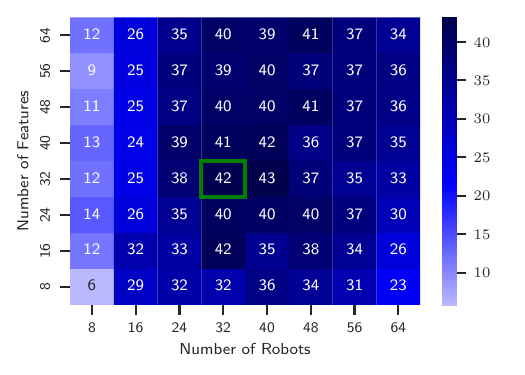}}\hspace{0.1cm}
    \subfloat{\includegraphics[width=0.45\textwidth]{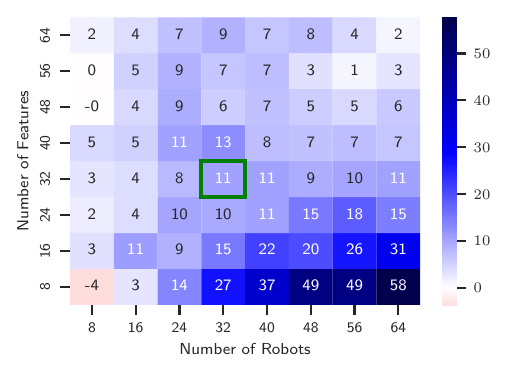}}\hspace{0.1cm}
    \caption{
        A heatmap showing the percent improvement in coverage cost of MAST-M compared to D-CVT (left) and LPAC-K3 (right) when the number of Gaussian features and agents is varied.
        Higher values indicate that MAST-M performs better than the baseline.
        The density that MAST-M and LPAC-K3 are trained on is highlighted in green.
        Each square shows the percentage improvement of MAST-M relative to the baseline.
        MAST-M outperforms D-CVT under all conditions and LPAC-K3 under most density scenarios.
        It struggles when the number of agents is small ($N = 8$) compared to its training data ($N = 32$).
        However, MAST-M is more robust to scenarios with low feature and high agent densities.
    }
    \label{fig:coverage-scale}
\end{figure*}

MAST-M and LPAC-K3 were trained in environments with $N=32$ agents and $F=32$ Gaussian features.
Regardless of these initial densities, agents may fail in the field, and the IDF may change.
Thus, a practical decentralized control policy should demonstrate transferability to varying densities of agents and features.
After training MAST-M and LPAC-K3 with $N=32$ agents and $F=32$ features, we evaluate the generalizability to out-of-distribution scenarios with varying density of agents and features.
We evaluate the policies for 64 combinations of $(N, F) \in \{8,16,24,32,40,48,56,64\}^2$.
The results are visualized in Figure~\ref{fig:coverage-scale} as heatmaps, which show the coverage cost improvement of MAST-M relative to D-CVT and LPAC-K3.
We compute the average normalized coverage cost over time and across 100 environments, then calculate the percentage improvement of MAST-M compared to D-CVT and LPAC-K3.

MAST-M performs better than D-CVT under all tested density combinations.
The smallest improvement over D-CVT is \SI{6}{\percent} which occurs at $N,F=(8, 8)$.
The largest improvement is \SI{43}{\percent} which occurs at $N,F=(40, 32)$.
The model, which was trained on $N, F=(32, 32)$, performs slightly better in this out-of-distribution case than in the in-distribution scenario.
The number of features in the IDF does not greatly impact the relative performance of MAST-M, but the number of robots does. 
MAST-M performance deteriorates when the number of agents is small.

MAST-M outperforms LPAC-K3 in most of the tested density scenarios.
In the worst case, MAST-M sees a \SI{4}{\percent} degradation in average coverage cost compared to LPAC-K3 at $N, F=(8,8)$.
However, MAST-M significantly outperforms LPAC-K3 in scenarios with a small number of features and a large number of robots.
In particular, MAST-M attains a \SI{58}{\percent} lower coverage cost than LPAC-K3 at $N,F=(64, 8)$.
We hypothesize that this may be an architectural limitation of GNNs, which struggle with modeling many-hop dependencies in the data.
When the feature-to-robot density ratio is low, information about peaks in the IDF needs to travel large distances.

\subsubsection{Real-World Environments}\label{subsec:coverage_real_world}

\begin{figure*}[htbp]
    \centering
    \rotatebox{90}{Philadelphia}
    \subfloat{\includegraphics[width=0.23\textwidth]{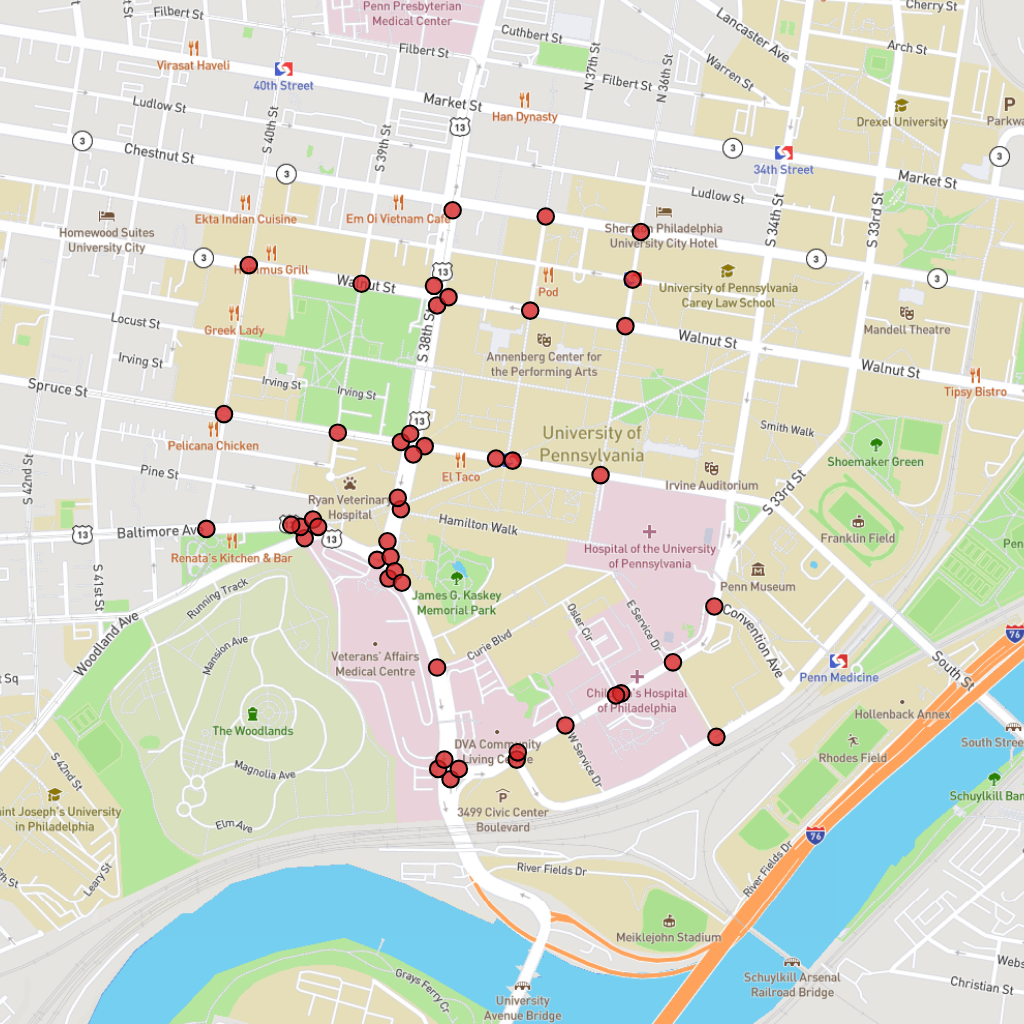}}\hspace{0.1cm}
    \subfloat{\includegraphics[width=0.23\textwidth, trim={70px 59px 29px 40px}, clip]{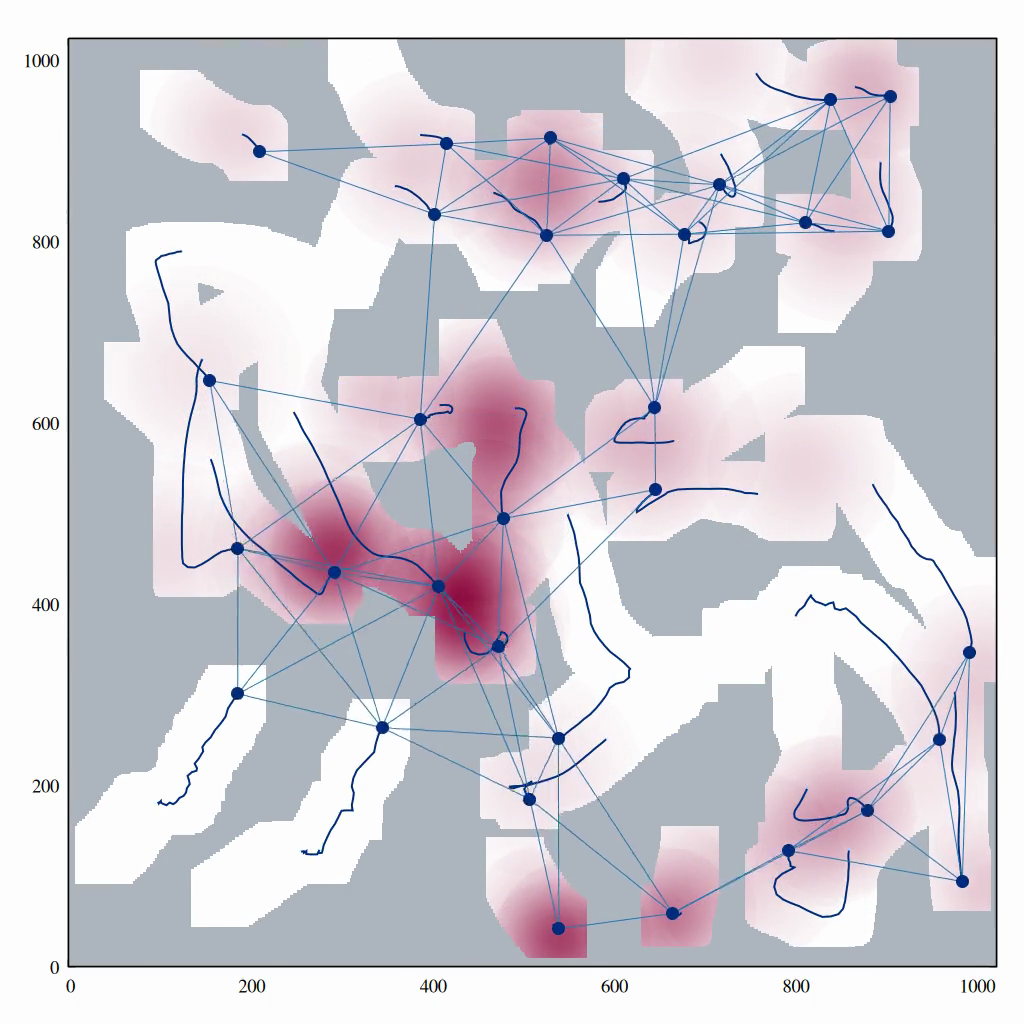}}\hspace{0.1cm}
    \subfloat{\includegraphics[width=0.23\textwidth, trim={70px 59px 29px 40px}, clip]{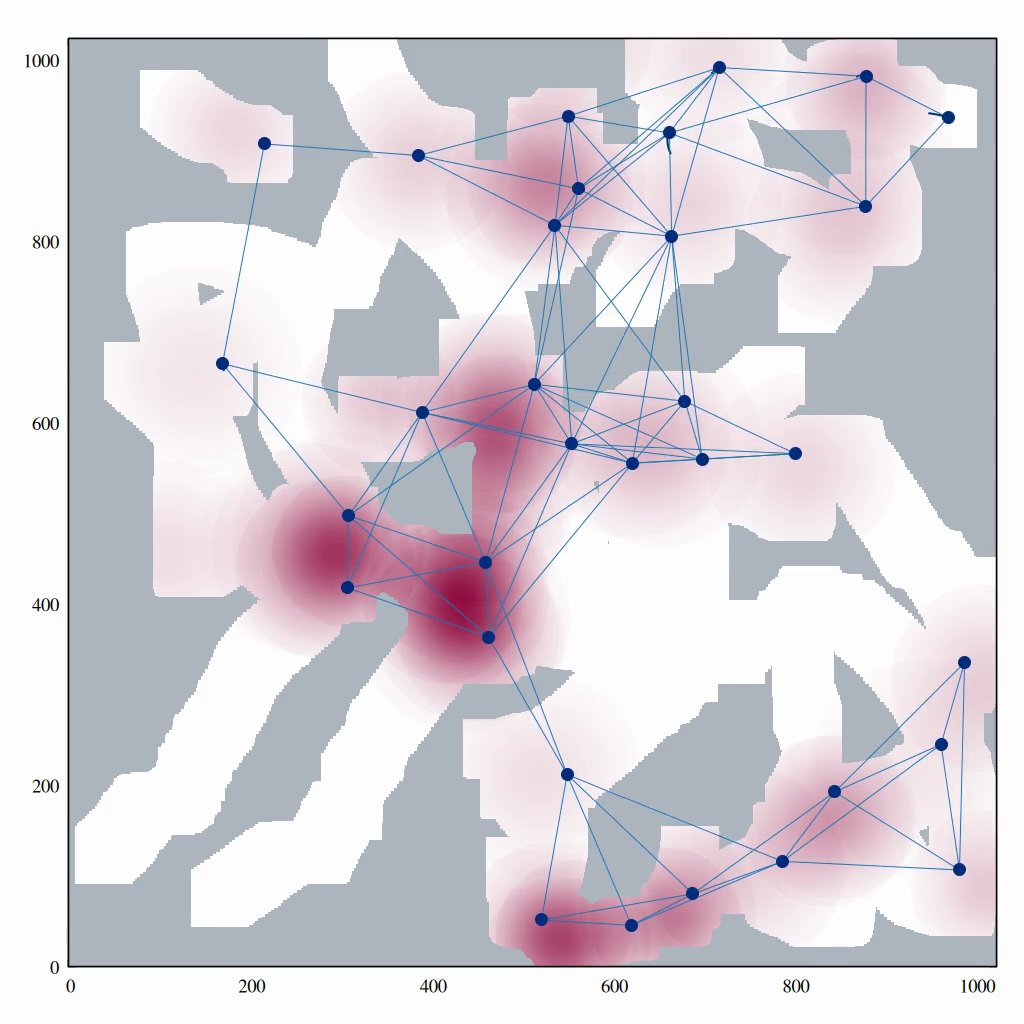}}\hspace{0.1cm}
    \subfloat{\includegraphics[width=0.23\textwidth, trim={70px 59px 29px 40px}, clip]{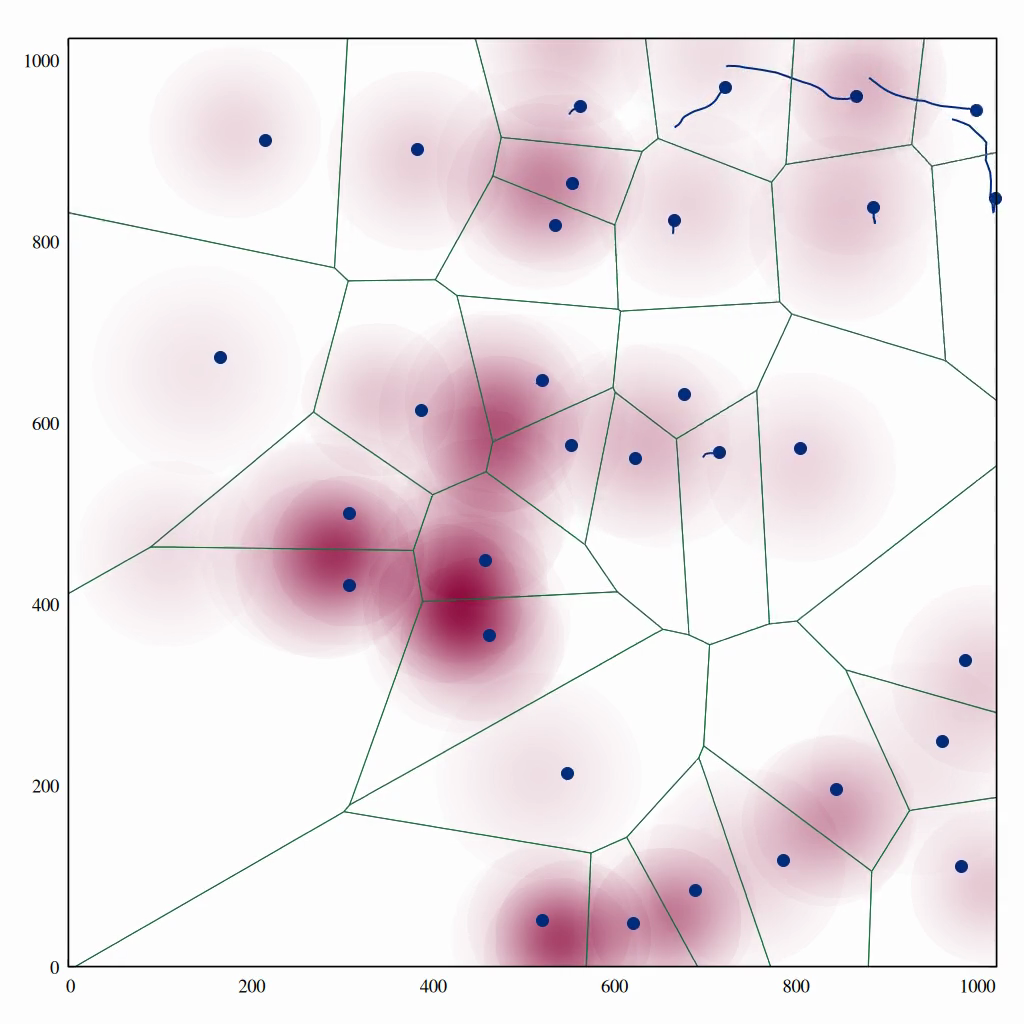}}
    \\
    \rotatebox{90}{Portland}
    \subfloat{\includegraphics[width=0.23\textwidth]{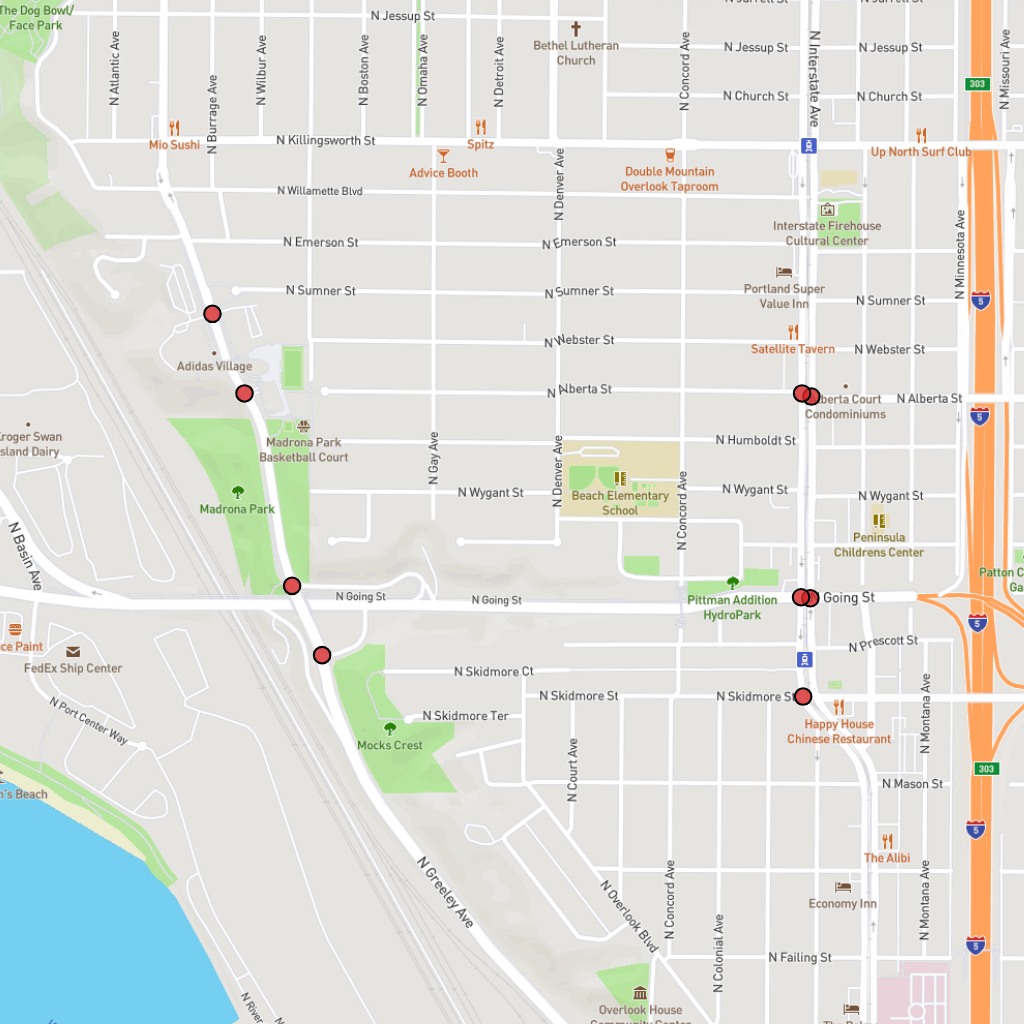}}\hspace{0.1cm}
    \subfloat{\includegraphics[width=0.23\textwidth, trim={70px 59px 29px 40px}, clip]{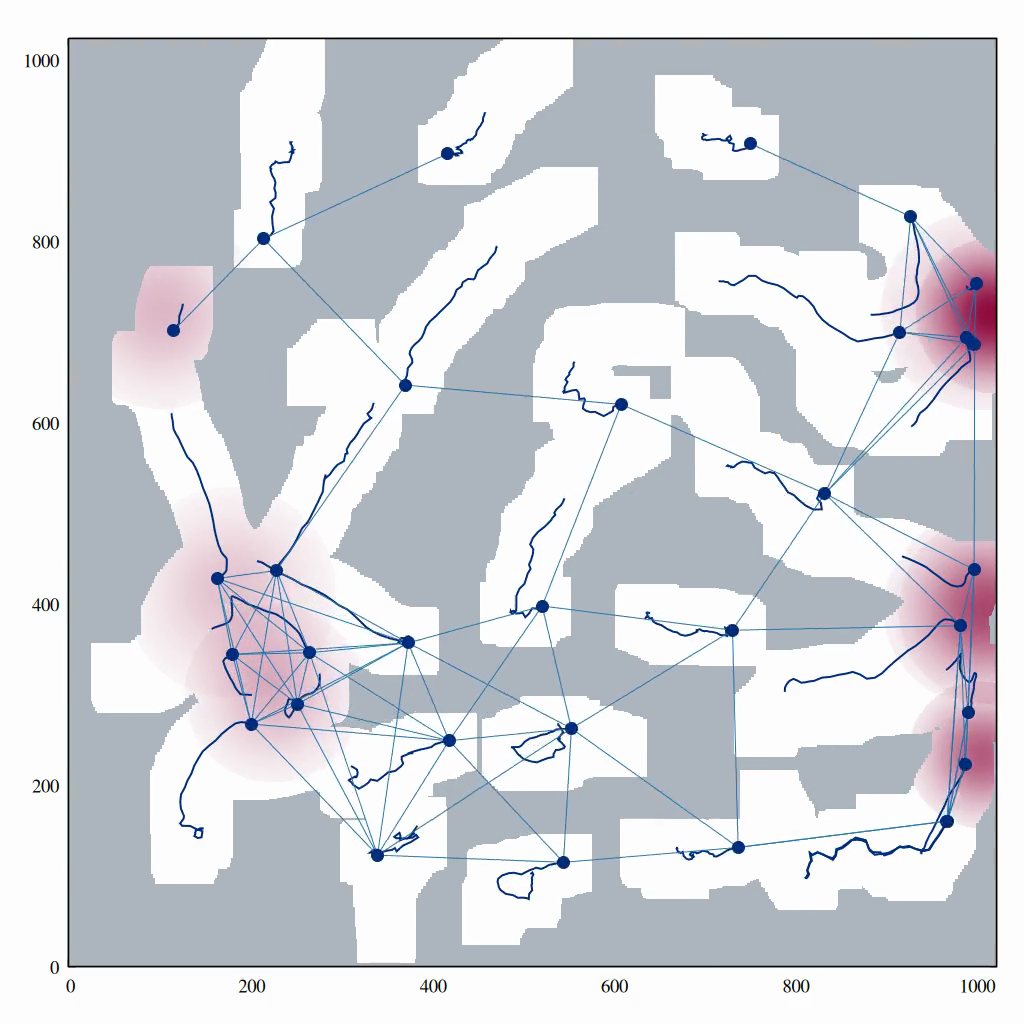}}\hspace{0.1cm}
    \subfloat{\includegraphics[width=0.23\textwidth, trim={70px 59px 29px 40px}, clip]{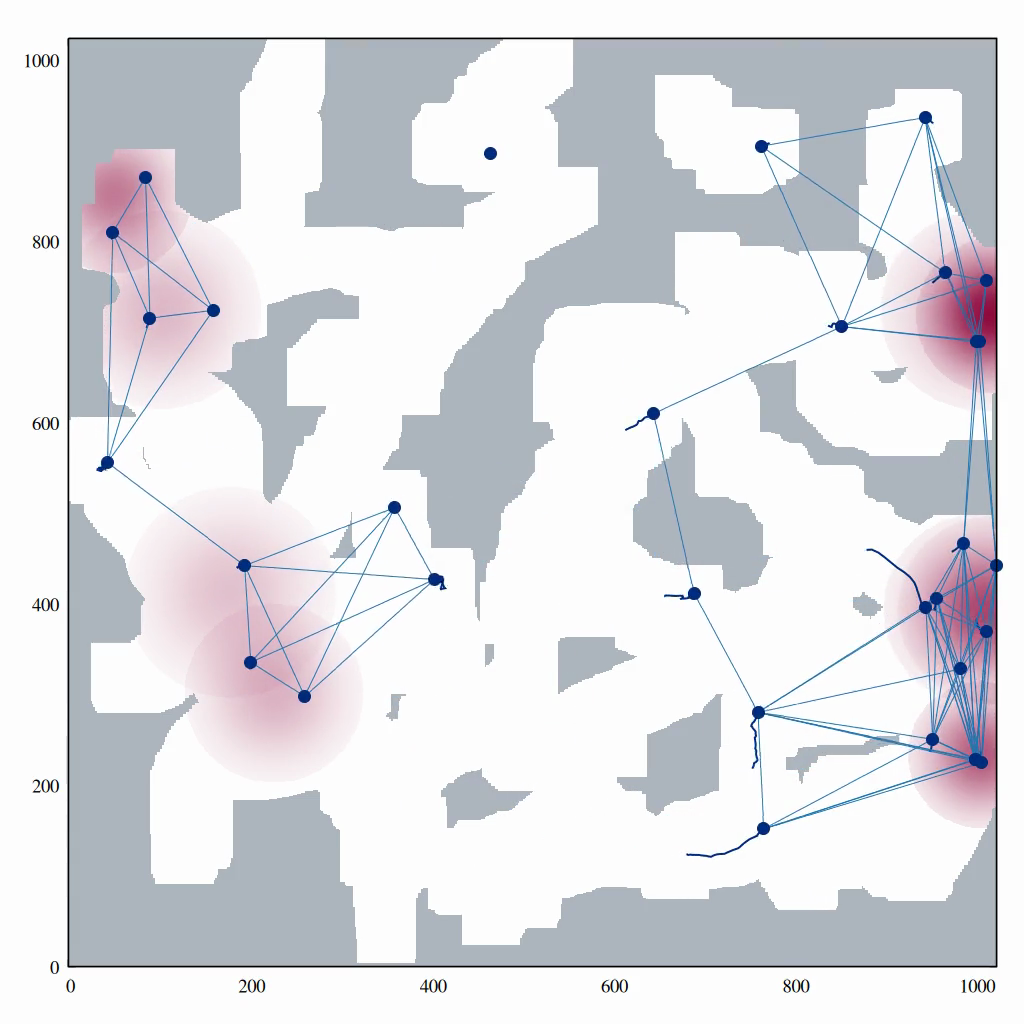}}\hspace{0.1cm}
    \subfloat{\includegraphics[width=0.23\textwidth, trim={70px 59px 29px 40px}, clip]{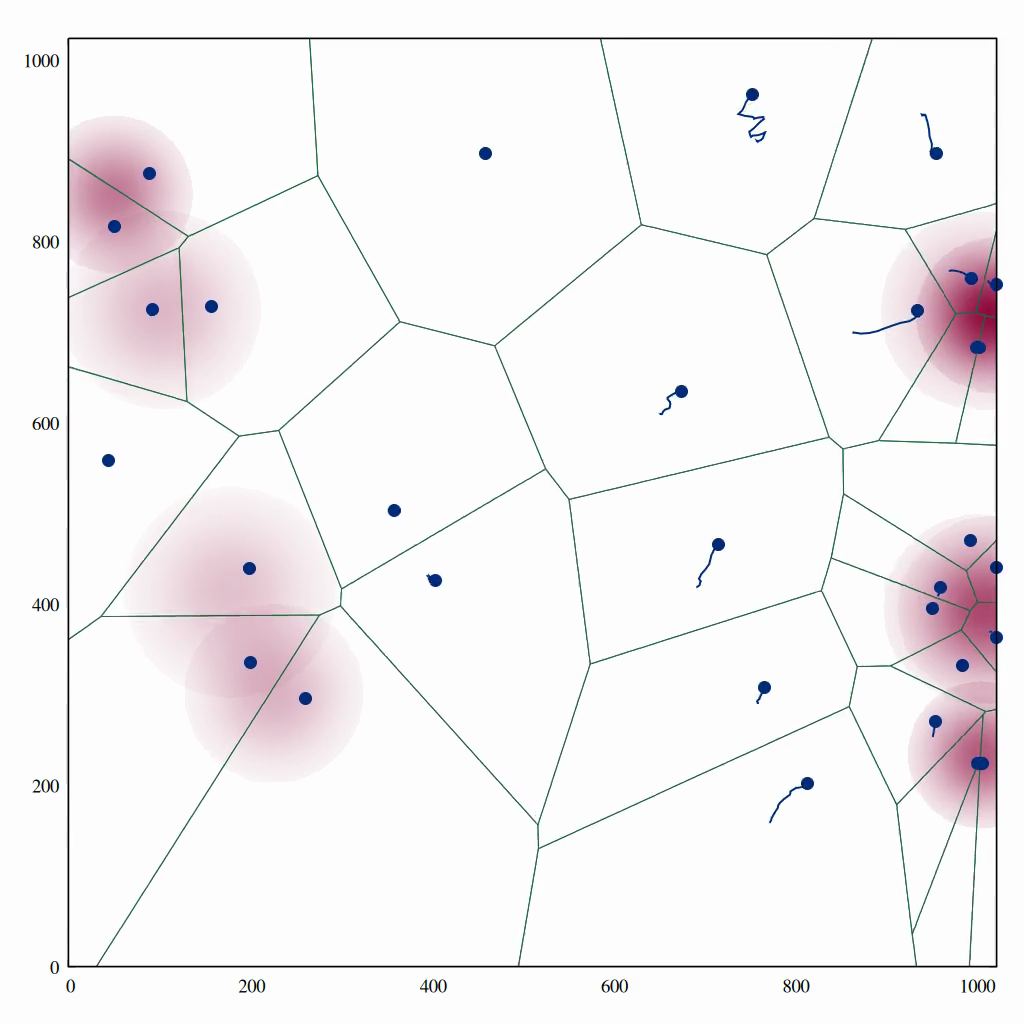}}
    \caption{
        Rollout of MAST-M on the Philadelphia and Portland environments with traffic light-based IDF.
        City maps are shown in the first column, with traffic-light locations in red.
        Columns two and three show the state of the multi-agent system after 100 and 500 steps of the MAST policy.
        The agent positions and communication graph are shown in blue, the IDF is shown in red,  and regions not observed by any agent are gray.
        The final column visualized the state after 600 steps, the complete IDF, and the Vornoi cells from Equation~\eqref{eq:voronoi_partition}.
    }
    \vspace{-0.5cm}
    \label{fig:coverage-real-world}
\end{figure*}

While randomly generated environments provide insight into model performance, we also evaluate MAST-M on environments informed by the real world.
We construct an IDF from the locations of traffic lights in $\qty{1024}{m} \times \qty{1024}{m}$ regions in 5 different cities, following \cite{Agarwal2024}.
The cities cover a broad spectrum of feature densities, with the number of features varying between 6 and 47, but the number of agents is fixed at $N = 32$.
In all scenarios, the initial positions of the agents are sampled from the uniform distribution.
To compare MAST-M and LPAC-K3, we initialize agents at the same positions.
The results of these experiments are reported in Table~\ref{tab:coverage-realworld}.
MAST-M outperforms LPAC-K3 in all 5 of these real-world scenarios.
Additionally, we visualize the MAST-M policy in Figure~\ref{fig:coverage-real-world} at $t \in \{100, 500, 600\}$ in the Philadelphia and Portland environments. 
In Philadelphia, agents spread out to cover the high-density environment, demonstrating that the team can balance exploration and coverage.
In Portland, the IDF is covered, although there is room for improvement since a few agents are left in areas with low IDF values.
This highlights a weakness in the dataset used for training: MAST-M and LPAC-K3 are overfitted to uniformly distributed constant-density environments.

\begin{table}
    \centering
    \begin{threeparttable}
        \caption{Average Normalized Coverage Cost on Traffic Light Environments}
        \label{tab:coverage-realworld}
        \begin{tabularx}{\tableWidth\linewidth}{YYYY}
            \toprule
            \textbf{Environment} & $F$ & \textbf{LPAC-K3 \cite{Agarwal2024}} & \textbf{MAST-M (Ours)} \\
            \midrule
            Philadelphia         & 47  & 0.492                               & \textbf{0.328}                  \\
            San Francisco        & 30  & 0.444                               & \textbf{0.404}                  \\
            Minneapolis          & 22  & 0.438                               & \textbf{0.386}                  \\
            Portland             & 9   & 0.195                               & \textbf{0.182}                  \\
            Seattle              & 6   & 0.324                               & \textbf{0.277}                  \\
            \bottomrule
        \end{tabularx}
        \begin{tablenotes}
            \item Recorded average normalized coverage cost for each environment.
            Comparison of MAST-M against LPAC-K3 for environments created from the traffic light dataset.
            MAST-M outperforms LPAC-K3 across all tested real-world scenarios.
            The number of features varies between cities, but we always use 32 agents.
        \end{tablenotes}
    \end{threeparttable}
\end{table}

\subsection{Discussion}

The results demonstrate that MAST-M outperforms the baselines on the coverage control task.
MAST-M also generalizes to out-of-distribution scenarios with varying agent and feature densities.
The generalizability properties of MAST-M on coverage control align with what we observe in DAN, despite the differences between the tasks.
Figure \ref{fig:coverage-scale} shows that MAST-M is much more transferable than LPAC-K3 in scenarios with many more agents than features.
We have reviewed videos of the two policies in those scenarios and observed that the LPAC-K3 policy is more biased towards maintaining a uniform separation between robots rather than adapting to the distribution of the IDF.
This behavior is less prominent with MAST-M, likely signifying that MAST-M can better capture long-distance relationships in the data, rather than relying on local heuristics.
Another reason for this difference in performance could be that LPAC-K3 comprises isotropic aggregations, which may make it challenging to communicate directional information.
This is not the case for MAST-M, which can leverage the positional encodings to learn anisotropic filters.
We believe that this gives MAST-M a significant advantage in most scenarios.

MAST-M also generalizes to larger-sized environments at constant agent and feature densities.
Across all environment sizes, it performs at least as well as LPAC-K3.
MAST-M has a significant advantage in smaller environments, but performs equally well to LPAC-K3 in the \SI{2048}{\meter}-wide environment.
Notice that LPAC-K3 has a much larger receptive field, aggregating information from a 15-hop neighborhood; the receptive field has a maximum radius of \SI{3,840}{\meter} since the communication radius is \SI{256}{\meter}.
On the other hand, MAST-M has only $L=8$ layers, each with a \SI{256}{\meter} wide receptive field for a total receptive field width of \SI{2,048}{\meter}.
LPAC-K3's larger receptive field gives it an advantage for larger-scale tasks, despite which it achieves equivalent performance, as shown in \fgref{fig:coverage-strip}.
However, the size-generalizability of MAST-M is imperfect, with its performance deteriorating as the scale of the task increases.


\appendices
\section{Other Positional Encodings}

\subsection{MLP Positional Encoding (MLP-PE)}\label{subsec:mlp-pe}
A simple way to include positional information is to learn an MLP that maps positions, $\bfp_i$, to a same-dimensional vector as the input, $\bfx_i$.
The MLP-PE is added to the input embedding:
\begin{equation}\label{eq:embedding_mlp}
    \bfz_i = \bfx_i + MLP(\bfp_i)
\end{equation}
where $\mlp \colon \reals^2 \mapsto \reals^d$ is a multi-layer perceptron (MLP). The embeddings $\bfz_i$ become inputs to self-attention instead of $\bfx_i$.
For models comprising multiple self-attention layers with residual connections, the MLP-PE is added only to the first layer.
Residual connections propagate the positional information through to subsequent layers.

\subsection{Absolute Positional Encoding (APE)}\label{subsec:ape}
The original transformer architecture~\cite{Vaswani2017} uses sinusoidal positional encodings, which are added to the input sequence.
Let $\APE(\bfp_i) \in \reals^d$ be a sinusoidal positional encoding for $\bfp_i$.
$\APE(\bfp_i)$ is generated using alternating sines and cosines with frequencies $\omega_k \in \reals_+$:
\begin{equation}\label{eq:absolute_positional_encodings}
    \APE(\bfp_i) =
    \begin{bmatrix}
        \vdots              \\
        \APE_{4k}(\bfp_i)   \\
        \APE_{4k+1}(\bfp_i) \\
        \APE_{4k+2}(\bfp_i) \\
        \APE_{4k+3}(\bfp_i) \\
        \vdots
    \end{bmatrix}
    = \begin{bmatrix}
        \vdots                   \\
        \sin(\omega_k  p_{i}^x ) \\
        \cos(\omega_k  p_{i}^x)  \\
        \sin(\omega_k  p_{i}^y)  \\
        \cos(\omega_k  p_{i}^y)  \\
        \vdots
    \end{bmatrix}
\end{equation}
Here, $k$ ranges from $1$ to $d/4$, $\APE_n(\bfp_i)$ is the $n$th element of $\APE(\bfp_i)$, and $p_i^x, p_i^y$ are the components of $\bfp_i$.
Similar to~\eqref{eq:embedding_ape}, the projection functions for APE have the form,
\begin{equation}\label{eq:embedding_ape}
    \bfz_i = \bfW_{\{q,k,v\}} \left( \bfx_i + \APE(\bfp_i) \right).
\end{equation}
As with MLP-PE, when there are multiple self-attention layers with residual connections, APE needs only to be added at the first layer.

\section*{Impact of attention windowing and positional encoding}\label{appendix:dan_gridserach}
\def\gridserachWidth{0.95}
\begin{figure*}[h]
    \centering
    \includegraphics[width=\gridserachWidth\linewidth]{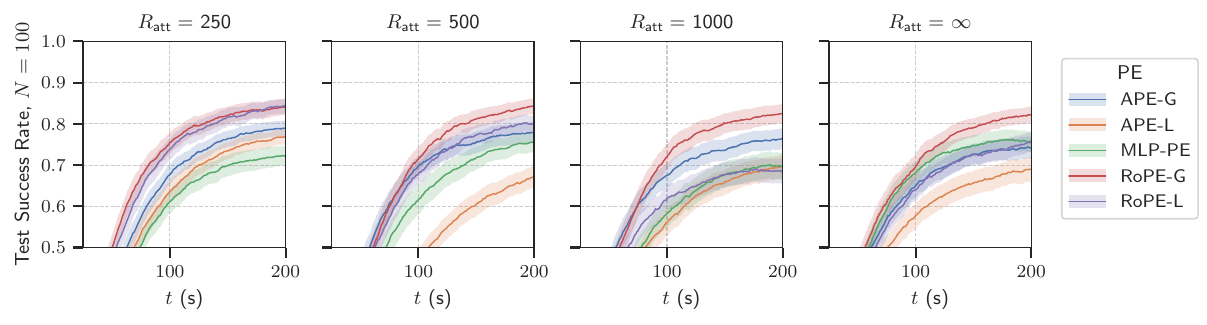}
    \includegraphics[width=\gridserachWidth\linewidth]{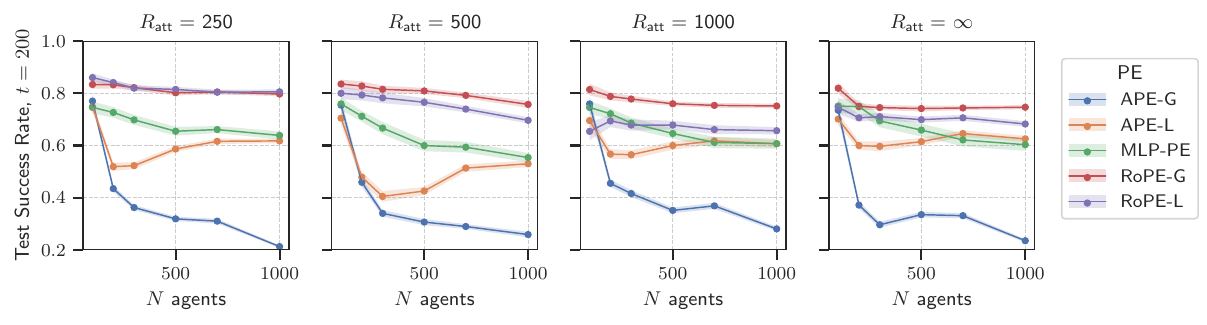}
    \caption{Mean DAN success rate obtained during testing after training with component masking (Section~\ref{subsec:centralized_training}). The model was trained with $N=100$ agents. The \emph{top row} shows the success rate over time for $N=100$ agents. The \emph{bottom row} shows the effect of the number of agents, $N \in \{100, 200, 300, 500, 700, 1000\}$, on the terminal success rate $SR(t = 200)$. Solid lines show the mean success rate across trials, while the shaded region indicates a 95\% confidence interval. The attention window radius $R_\mathrm{att}$ varies along the columns. Colors indicate the positional encoding.}
    \label{fig:masked-comparison}
\end{figure*}
\begin{figure*}[h]
    \centering
    \includegraphics[width=\gridserachWidth\linewidth]{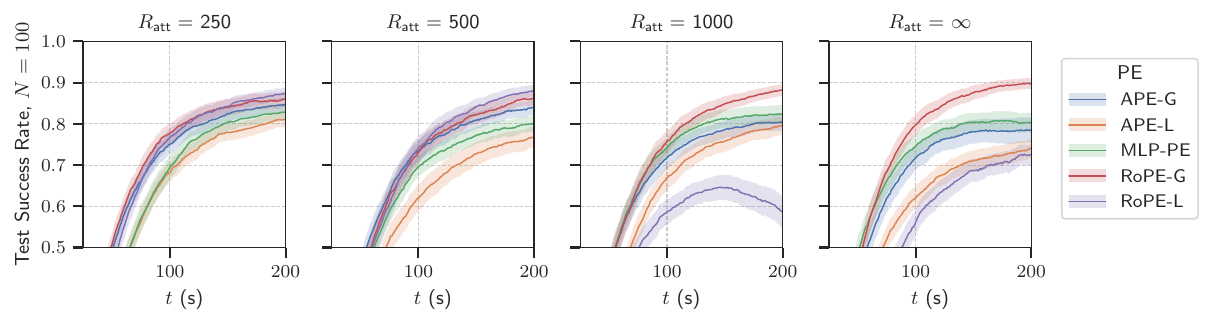}
    \includegraphics[width=\gridserachWidth\linewidth]{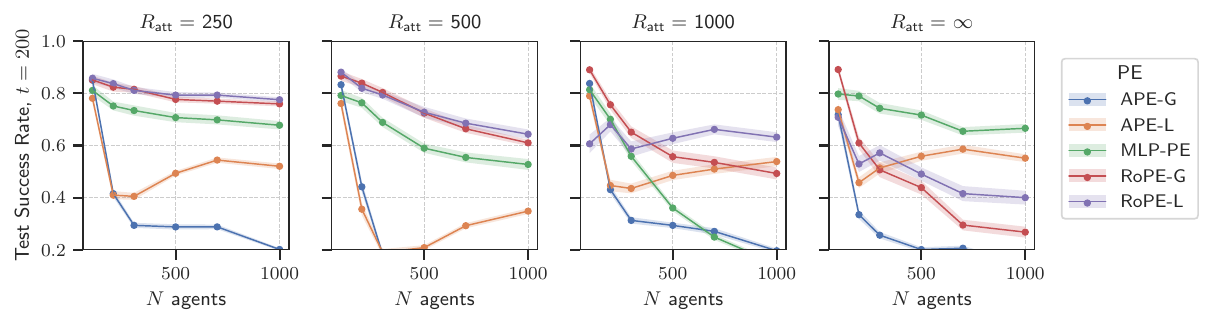}
    \caption{Mean DAN success rate obtained during testing after training without component masking (Section~\ref{subsec:centralized_training}). The model was trained with $N=100$ agents. The \emph{top row} shows the success rate over time for $N=100$ agents. The \emph{bottom row} shows the effect of the number of agents, $N \in \{100, 200, 300, 500, 700, 1000\}$, on the terminal success rate $SR(t = 200)$. Solid lines show the mean success rate across trials, while the shaded region indicates a 95\% confidence interval. The attention window radius $R_\mathrm{att}$ varies along the columns. Colors indicate the positional encoding.}
    \label{fig:local-comparison}
\end{figure*}

\bibliographystyle{IEEEtran}

\begin{thebibliography}{10}
\providecommand{\url}[1]{#1}
\csname url@samestyle\endcsname
\providecommand{\newblock}{\relax}
\providecommand{\bibinfo}[2]{#2}
\providecommand{\BIBentrySTDinterwordspacing}{\spaceskip=0pt\relax}
\providecommand{\BIBentryALTinterwordstretchfactor}{4}
\providecommand{\BIBentryALTinterwordspacing}{\spaceskip=\fontdimen2\font plus
\BIBentryALTinterwordstretchfactor\fontdimen3\font minus \fontdimen4\font\relax}
\providecommand{\BIBforeignlanguage}[2]{{%
\expandafter\ifx\csname l@#1\endcsname\relax
\typeout{** WARNING: IEEEtran.bst: No hyphenation pattern has been}%
\typeout{** loaded for the language `#1'. Using the pattern for}%
\typeout{** the default language instead.}%
\else
\language=\csname l@#1\endcsname
\fi
#2}}
\providecommand{\BIBdecl}{\relax}
\BIBdecl
\renewcommand{\BIBentryALTinterwordstretchfactor}{4}

\bibitem{Kegeleirs21-SwarmSLAM}
M.~Kegeleirs, G.~Grisetti, and M.~Birattari, ``Swarm {{SLAM}}: {{Challenges}} and {{Perspectives}},'' \emph{Front Robot AI}, vol.~8, p. 618268, Mar. 2021.

\bibitem{Lu25-Reinforcement}
T.~Lu, D.~Sobti, D.~Talwar, and W.~Wu, ``Reinforcement learning-based dynamic field exploration and reconstruction using multi-robot systems for environmental monitoring,'' \emph{Front. Robot. AI}, vol.~12, p. 1492526, Mar. 2025.

\bibitem{Cao24-HMASARMultiAgentSearch}
X.~Cao, M.~Li, Y.~Tao, and P.~Lu, ``{{HMA-SAR}}: {{Multi-Agent Search}} and {{Rescue}} for {{Unknown Located Dynamic Targets}} in {{Completely Unknown Environments}},'' \emph{IEEE Robotics and Automation Letters}, vol.~9, no.~6, pp. 5567--5574, Jun. 2024.

\bibitem{Bi24-CUREHierarchical}
Q.~Bi \emph{et~al.}, ``{{CURE}}: {{A Hierarchical Framework}} for {{Multi-Robot Autonomous Exploration Inspired}} by {{Centroids}} of {{Unknown Regions}},'' \emph{IEEE Transactions on Automation Science and Engineering}, vol.~21, no.~3, pp. 3773--3786, Jul. 2024.

\bibitem{Agarwal2024}
S.~Agarwal, R.~Muthukrishnan, W.~Gosrich, A.~Ribeiro, and V.~Kumar, ``{{LPAC}}: {{Learnable}} perception-action-communication loops with applications to coverage control,'' \emph{arXiv preprint arXiv:2401.04855}, p. 2401.04855, 2024.

\bibitem{Hu22-Scalable}
T.-K. Hu \emph{et~al.}, ``Scalable {{Perception-Action-Communication Loops With Convolutional}} and {{Graph Neural Networks}},'' \emph{IEEE Transactions on Signal and Information Processing over Networks}, vol.~8, pp. 12--24, 2022.

\bibitem{Li2020}
Q.~Li, F.~Gama, A.~Ribeiro, and A.~Prorok, ``Graph neural networks for decentralized multi-robot path planning,'' in \emph{2020 {{IEEE}}/{{RSJ}} International Conference on Intelligent Robots and Systems ({{IROS}})}, 2020, pp. 11\,785--11\,792.

\bibitem{Tolstaya20-Learning}
E.~Tolstaya, F.~Gama, J.~Paulos, G.~Pappas, V.~Kumar, and A.~Ribeiro, ``Learning decentralized controllers for robot swarms with graph neural networks,'' in \emph{Proceedings of the Conference on Robot Learning}, ser. Proceedings of Machine Learning Research, vol. 100.\hskip 1em plus 0.5em minus 0.4em\relax PMLR, 2020-10-30/2020-11-01, pp. 671--682.

\bibitem{Zhou2022}
L.~Zhou \emph{et~al.}, ``Graph neural networks for decentralized multi-robot target tracking,'' in \emph{2022 {{IEEE}} International Symposium on Safety, Security, and Rescue Robotics ({{SSRR}})}, 2022, pp. 195--202.

\bibitem{Gosrich2022}
W.~Gosrich \emph{et~al.}, ``Coverage control in multi-robot systems via graph neural networks,'' in \emph{2022 International Conference on Robotics and Automation ({{ICRA}})}, 2022, pp. 8787--8793.

\bibitem{Ruiz21-GraphNeuralNetworks}
L.~Ruiz, F.~Gama, and A.~Ribeiro, ``Graph {{Neural Networks}}: {{Architectures}}, {{Stability}}, and {{Transferability}},'' \emph{Proceedings of the IEEE}, vol. 109, no.~5, pp. 660--682, May 2021.

\bibitem{Ruiz20-GraphonNeural}
L.~Ruiz, L.~Chamon, and A.~Ribeiro, ``Graphon neural networks and the transferability of graph neural networks,'' \emph{Advances in Neural Information Processing Systems}, vol.~33, pp. 1702--1712, 2020.

\bibitem{Alon20-BottleneckGraphNeural}
U.~Alon and E.~Yahav, ``On the {{Bottleneck}} of {{Graph Neural Networks}} and its {{Practical Implications}},'' in \emph{International {{Conference}} on {{Learning Representations}}}, Oct. 2020.

\bibitem{Topping21-Understanding}
J.~Topping, F.~D. Giovanni, B.~P. Chamberlain, X.~Dong, and M.~M. Bronstein, ``Understanding over-squashing and bottlenecks on graphs via curvature,'' in \emph{International {{Conference}} on {{Learning Representations}}}, Oct. 2021.

\bibitem{Chen23-TransformerBased}
L.~Chen \emph{et~al.}, ``Transformer-{{Based Imitative Reinforcement Learning}} for {{Multirobot Path Planning}},'' \emph{IEEE Transactions on Industrial Informatics}, vol.~19, no.~10, pp. 10\,233--10\,243, Oct. 2023.

\bibitem{Chen24-TransformerBased}
Q.~Chen, R.~Wang, M.~Lyu, and J.~Zhang, ``Transformer-{{Based Reinforcement Learning}} for {{Multi-Robot Autonomous Exploration}},'' \emph{Sensors}, vol.~24, no.~16, p. 5083, Jan. 2024.

\bibitem{Wang23-NaviSTARSociallyAware}
W.~Wang, R.~Wang, L.~Mao, and B.-C. Min, ``{{NaviSTAR}}: {{Socially Aware Robot Navigation}} with {{Hybrid Spatio-Temporal Graph Transformer}} and {{Preference Learning}},'' in \emph{2023 {{IEEE}}/{{RSJ International Conference}} on {{Intelligent Robots}} and {{Systems}} ({{IROS}})}, Oct. 2023, pp. 11\,348--11\,355.

\bibitem{Yuan21-AgentFormerAgentAware}
Y.~Yuan, X.~Weng, Y.~Ou, and K.~Kitani, ``{{AgentFormer}}: {{Agent-Aware Transformers}} for {{Socio-Temporal Multi-Agent Forecasting}},'' in \emph{2021 {{IEEE}}/{{CVF International Conference}} on {{Computer Vision}} ({{ICCV}})}.\hskip 1em plus 0.5em minus 0.4em\relax Montreal, QC, Canada: IEEE, Oct. 2021, pp. 9793--9803.

\bibitem{Zhang24-Decentralized}
Y.~Zhang, C.~Bai, B.~Zhao, J.~Yan, X.~Li, and X.~Li, ``Decentralized {{Transformers}} with {{Centralized Aggregation}} are {{Sample-Efficient Multi-Agent World Models}},'' p. 2406.15836, Jun. 2024.

\bibitem{Jaegle21-PerceiverGeneral}
A.~Jaegle, F.~Gimeno, A.~Brock, O.~Vinyals, A.~Zisserman, and J.~Carreira, ``Perceiver: {{General Perception}} with {{Iterative Attention}},'' in \emph{Proceedings of the 38th {{International Conference}} on {{Machine Learning}}}.\hskip 1em plus 0.5em minus 0.4em\relax PMLR, Jul. 2021, pp. 4651--4664.

\bibitem{Wen22-MultiagentReinforcement}
M.~Wen \emph{et~al.}, ``Multi-agent reinforcement learning is a sequence modeling problem,'' in \emph{Proceedings of the 36th {{International Conference}} on {{Neural Information Processing Systems}}}, ser. {{NIPS}} '22.\hskip 1em plus 0.5em minus 0.4em\relax Red Hook, NY, USA: Curran Associates Inc., Nov. 2022, pp. 16\,509--16\,521.

\bibitem{Egorov22-ScalableMultiAgent}
V.~Egorov and A.~Shpilman, ``Scalable {{Multi-Agent Model-Based Reinforcement Learning}},'' p. 2205.15023, May 2022.

\bibitem{Farjadnasab25-Cooperative}
M.~Farjadnasab and S.~Sirouspour, ``Cooperative and {{Asynchronous Transformer-based Mission Planning}} for {{Heterogeneous Teams}} of {{Mobile Robots}},'' p. 2410.06372, Jan. 2025.

\bibitem{Vaswani2017}
A.~Vaswani \emph{et~al.}, ``Attention is all you need,'' in \emph{Advances in Neural Information Processing Systems}, I.~Guyon \emph{et~al.}, Eds., vol.~30.\hskip 1em plus 0.5em minus 0.4em\relax Curran Associates, Inc., 2017.

\bibitem{Bronstein21-GeometricDeep}
M.~M. Bronstein, J.~Bruna, T.~Cohen, and P.~Veli{\v c}kovi{\'c}, ``Geometric {{Deep Learning}}: {{Grids}}, {{Groups}}, {{Graphs}}, {{Geodesics}}, and {{Gauges}},'' p. 2104.13478, May 2021.

\bibitem{Zhao21-PointTransformer}
H.~Zhao, L.~Jiang, J.~Jia, P.~Torr, and V.~Koltun, ``Point {{Transformer}},'' p. 2012.09164, Sep. 2021.

\bibitem{Su2024}
J.~Su, M.~Ahmed, Y.~Lu, S.~Pan, W.~Bo, and Y.~Liu, ``{{RoFormer}}: {{Enhanced}} transformer with rotary position embedding,'' \emph{Neurocomputing}, vol. 568, p. 127063, Feb. 2024.

\bibitem{Heo2024}
B.~Heo, S.~Park, D.~Han, and S.~Yun, ``Rotary position embedding for vision transformer,'' in \emph{Computer Vision - {{ECCV}} 2024}.\hskip 1em plus 0.5em minus 0.4em\relax Springer Nature Switzerland, Nov. 2024, pp. 289--305.

\bibitem{Klavins04-Communication}
E.~Klavins, ``Communication {{Complexity}} of {{Multi-robot Systems}},'' in \emph{Algorithmic {{Foundations}} of {{Robotics V}}}, J.-D. Boissonnat, J.~Burdick, K.~Goldberg, and S.~Hutchinson, Eds.\hskip 1em plus 0.5em minus 0.4em\relax Berlin, Heidelberg: Springer, 2004, pp. 275--291.

\bibitem{Khan2021}
A.~Khan, V.~Kumar, and A.~Ribeiro, ``Large scale distributed collaborative unlabeled motion planning with graph policy gradients,'' \emph{IEEE Robot. Autom. Lett.}, vol.~6, no.~3, pp. 5340--5347, Jul. 2021.

\bibitem{Xiong20-LayerNormalization}
R.~Xiong \emph{et~al.}, ``On layer normalization in the transformer architecture,'' in \emph{Proceedings of the 37th International Conference on Machine Learning}, ser. {{ICML}}'20.\hskip 1em plus 0.5em minus 0.4em\relax JMLR.org, 2020.

\bibitem{Owerko23-Transferability}
D.~Owerko, C.~I. Kanatsoulis, J.~Bondarchuk, D.~J. Bucci~Jr, and A.~Ribeiro, ``Transferability of {{Convolutional Neural Networks}} in {{Stationary Learning Tasks}},'' Jul. 2023.

\bibitem{Mox24-Opportunistic}
D.~Mox, K.~Garg, A.~Ribeiro, and V.~Kumar, ``Opportunistic {{Communication}} in {{Robot Teams}},'' in \emph{2024 {{IEEE International Conference}} on {{Robotics}} and {{Automation}} ({{ICRA}})}, May 2024, pp. 12\,090--12\,096.

\bibitem{Turpin2014}
M.~Turpin, N.~Michael, and V.~Kumar, ``{{CAPT}}: {{Concurrent}} assignment and planning of trajectories for multiple robots,'' \emph{International Journal of Robotics Research}, vol.~33, no.~1, pp. 98--112, Jan. 2014.

\bibitem{Panagou2014}
D.~Panagou, M.~Turpin, and V.~Kumar, ``Decentralized goal assignment and trajectory generation in multi-robot networks: {{A}} multiple lyapunov functions approach,'' in \emph{2014 {{IEEE}} International Conference on Robotics and Automation ({{ICRA}})}.\hskip 1em plus 0.5em minus 0.4em\relax IEEE, 2014, pp. 6757--6762.

\bibitem{Goarin2024}
M.~Goarin and G.~Loianno, ``Graph neural network for decentralized multi-robot goal assignment,'' \emph{IEEE Robotics and Automation Letters}, vol.~9, no.~5, pp. 4051--4058, May 2024.

\bibitem{Zou05-RegularizationVariable}
H.~Zou and T.~Hastie, ``Regularization and {{Variable Selection Via}} the {{Elastic Net}},'' \emph{Journal of the Royal Statistical Society Series B: Statistical Methodology}, vol.~67, no.~2, pp. 301--320, Apr. 2005.

\bibitem{Kuhn1955}
H.~W. Kuhn, ``The {{Hungarian}} method for the assignment problem,'' \emph{Naval research logistics quarterly}, vol.~2, no. 1-2, pp. 83--97, 1955.

\bibitem{Ismail2017}
S.~Ismail and L.~Sun, ``Decentralized hungarian-based approach for fast and scalable task allocation,'' in \emph{2017 International Conference on Unmanned Aircraft Systems ({{ICUAS}})}.\hskip 1em plus 0.5em minus 0.4em\relax IEEE, 2017, pp. 23--28.

\bibitem{Lam15-NumbaLLVMbasedPython}
S.~K. Lam, A.~Pitrou, and S.~Seibert, ``Numba: A {{LLVM-based Python JIT}} compiler,'' in \emph{Proceedings of the {{Second Workshop}} on the {{LLVM Compiler Infrastructure}} in {{HPC}}}.\hskip 1em plus 0.5em minus 0.4em\relax Austin Texas: ACM, Nov. 2015, pp. 1--6.

\bibitem{Bou23-TorchRLDatadriven}
A.~Bou \emph{et~al.}, ``{{TorchRL}}: {{A}} data-driven decision-making library for {{PyTorch}},'' 2023.

\bibitem{Loshchilov2017}
I.~Loshchilov, ``Decoupled weight decay regularization,'' \emph{arXiv preprint arXiv:1711.05101}, p. 1711.05101, 2017.

\bibitem{Li17-HyperbandNovel}
L.~Li, K.~Jamieson, G.~DeSalvo, A.~Rostamizadeh, and A.~Talwalkar, ``Hyperband: A novel bandit-based approach to hyperparameter optimization,'' \emph{J. Mach. Learn. Res.}, vol.~18, no.~1, pp. 6765--6816, Jan. 2017.

\bibitem{Bergstra11-Algorithms}
J.~Bergstra, R.~Bardenet, Y.~Bengio, and B.~K{\'e}gl, ``Algorithms for {{Hyper-Parameter Optimization}},'' in \emph{Advances in {{Neural Information Processing Systems}}}, vol.~24.\hskip 1em plus 0.5em minus 0.4em\relax Curran Associates, Inc., 2011.

\bibitem{Akiba19-OptunaNextgeneration}
T.~Akiba, S.~Sano, T.~Yanase, T.~Ohta, and M.~Koyama, ``Optuna: {{A Next-generation Hyperparameter Optimization Framework}},'' p. 1907.10902, Jul. 2019.

\bibitem{CortesMKB02}
J.~Cort{\'e}s, S.~Martinez, T.~Karatas, and F.~Bullo, ``Coverage control for mobile sensing networks: {{Variations}} on a theme,'' in \emph{Mediterranean Conference on Control and Automation}.\hskip 1em plus 0.5em minus 0.4em\relax Lisbon, Portugal Lisbon, Portugal, 2002, pp. 9--13.

\bibitem{Lloyd82-LeastSquares}
S.~Lloyd, ``Least squares quantization in {{PCM}},'' \emph{IEEE Transactions on Information Theory}, vol.~28, no.~2, pp. 129--137, Mar. 1982.

\bibitem{Cortes05-Spatiallydistributed}
J.~Cort{\'e}s, S.~Mart{\'i}nez, and F.~Bullo, ``Spatially-distributed coverage optimization and control with limited-range interactions,'' \emph{ESAIM: COCV}, vol.~11, no.~4, pp. 691--719, Oct. 2005.

\bibitem{Rudolph21-RangeLimited}
M.~Rudolph, S.~Wilson, and M.~Egerstedt, ``Range {{Limited Coverage Control}} using {{Air-Ground Multi-Robot Teams}},'' in \emph{2021 {{IEEE International Conference}} on {{Robotics}} and {{Automation}} ({{ICRA}})}, May 2021, pp. 3525--3530.

\bibitem{Bai22-AdaptiveMultiAgent}
Y.~Bai, Y.~Wang, M.~Svinin, E.~Magid, and R.~Sun, ``Adaptive {{Multi-Agent Coverage Control With Obstacle Avoidance}},'' \emph{IEEE Control Systems Letters}, vol.~6, pp. 944--949, 2022.

\bibitem{Carron20-ModelPredictive}
A.~Carron and M.~N. Zeilinger, ``Model {{Predictive Coverage Control}}⁎,'' \emph{IFAC-PapersOnLine}, vol.~53, no.~2, pp. 6107--6112, Jan. 2020.

\end{thebibliography}

\end{document}